\documentclass[runningheads]{llncs}

\usepackage{graphicx}
\usepackage{comment}
\usepackage{multirow}
\usepackage{multicol}
\usepackage{amsmath,amssymb}
\usepackage{color}
\usepackage{url}
\usepackage{hyperref}
\usepackage{float}
\usepackage{caption,subcaption}
\usepackage{placeins}

\newif\ifreview
\reviewfalse

\ifreview
	\usepackage{lineno}

	\linenumbers
\fi

\begin{document}


\def\SubNumber{101}

\def\GCPRTrack{Fast Review Track}

\title{From Pointwise to Powerhouse: Initialising Neural Networks with Generative Models}

\ifreview
\else
	\titlerunning{From Pointwise to Powerhouse}

	\author{Christian Harder\inst{1,2}\orcidID{0000-0002-2953-8731} \and
	Moritz Fuchs\inst{1}\orcidID{0000-0003-3496-7271} \and
    Yuri Tolkach \inst{2}\orcidID{0000-0001-5239-2841} \and
	Anirban Mukhopadhyay\inst{1}\orcidID{0000-0003-0669-4018}
    }
	
	\authorrunning{C. Harder et al.}
	
	\institute{Technical University of Darmstadt, 64289 Darmstadt, Germany \and 
                University Hospital Cologne, 50937 Cologne, Germany}
\fi

\maketitle              

\begin{abstract}
Traditional initialisation methods, e.g. He and Xavier, have been effective in avoiding the problem of vanishing or exploding gradients in neural networks. However, they only use simple pointwise distributions, which model one-dimensional variables. Moreover, they ignore most information about the architecture and disregard past training experiences. These limitations can be overcome by employing generative models for initialisation.  
\newline
In this paper, we introduce two groups of new initialisation methods. First, we \textit{locally} initialise weight groups by employing variational autoencoders. Secondly, we \textit{globally} initialise full weight sets by employing graph hypernetworks. We thoroughly evaluate the impact of the employed generative models on state-of-the-art neural networks in terms of accuracy, convergence speed and ensembling. Our results show that \textit{global} initialisations result in higher accuracy and faster initial convergence speed. However, the implementation through graph hypernetworks leads to diminished ensemble performance on out of distribution data. To counteract, we propose a modification called \textit{noise graph hypernetwork}, which encourages diversity in the produced ensemble members. Furthermore, our approach might be able to transfer learned knowledge to different image distributions. Our work provides insights into the potential, the trade-offs and possible modifications of these new initialisation methods.
\end{abstract}
\section{Introduction}
Neural networks have shown remarkable success in various computer vision tasks, such as image classification \cite{ciregan2012multi,he2015delving,russakovsky2015imagenet,wan2013regularization}, segmentation \cite{cao2023swin,huang2020unet,strudel2021segmenter}, and detection \cite{girshick2014rich,ren2015faster}. Their performance depends on several factors, including their architecture, quality of data and available computing resources. Among these factors, the \textbf{choice of weight initialisation} plays a crucial role in the network's performance \cite{sutskever2013importance}. A proper initialisation greatly enhances training convergence, whereas a poor one can hinder it \cite{he2015delving}.
These findings have sparked entire fields of research dedicated to exploring innovative approaches. Techniques such as self-supervised learning \cite{jing2020self}, knowledge distillation \cite{gou2021knowledge}, and transfer learning \cite{tan2018survey} have emerged as notable approaches. In contrast, our work focuses on initialisations without training during the weight generation.\\
\begin{figure}[H]
    \centering
    \includegraphics[width=0.6\linewidth]{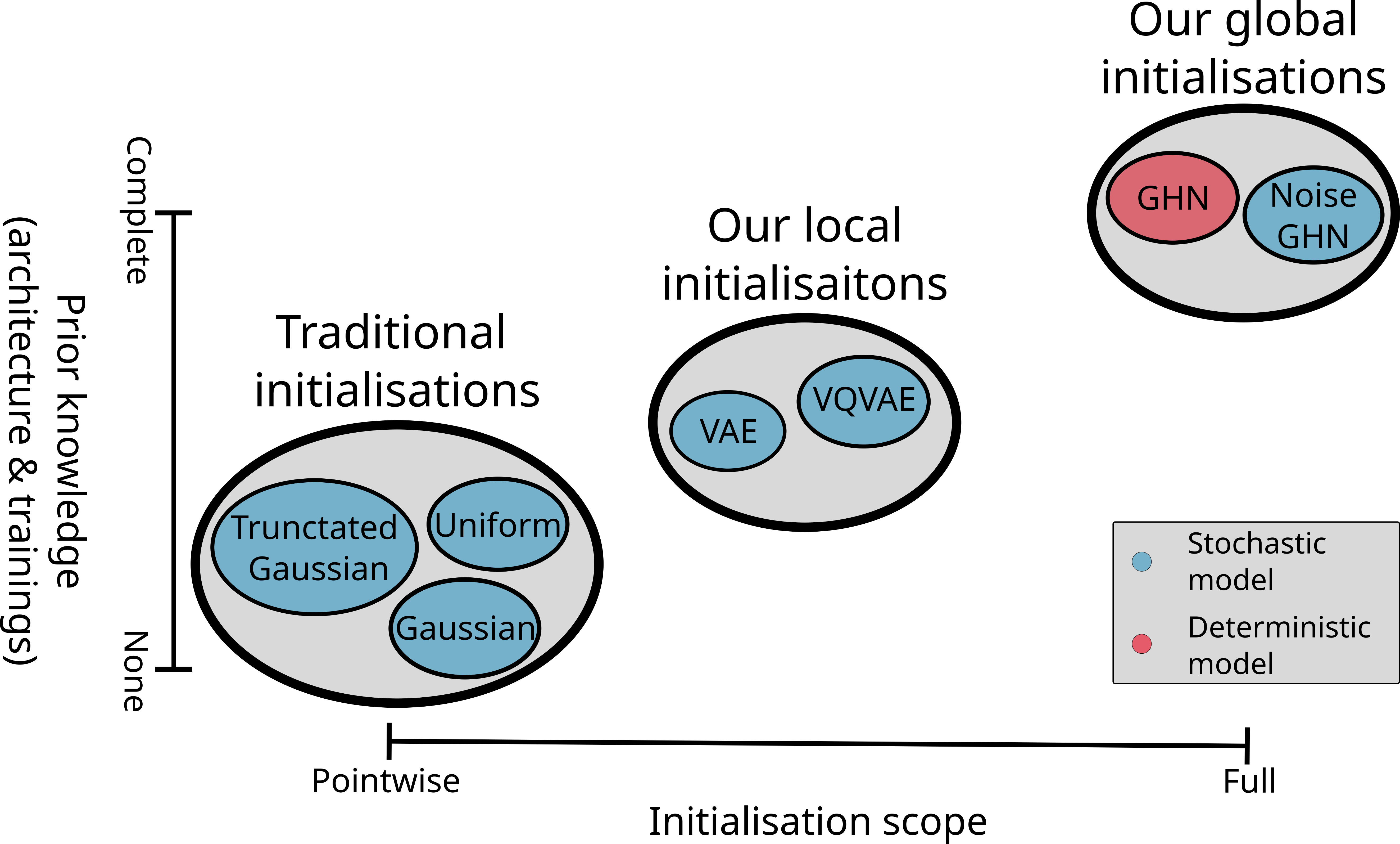}
    \caption{Traditional initialisation methods only consider layer dimensions, while being simple pointwise distributions. Utilising generative models, we consider significantly more architecture information and model complete weight sets by employing graph hypernetworks (GHN).}
    \label{fig:approach}
\end{figure}
\noindent Traditional initialisation methods such as He \cite{he2015delving} and Xavier \cite{glorot2010understanding} initialisation were designed to avoid the problem of vanishing or exploding gradients. Despite their effectiveness, these approaches have \textbf{two significant weaknesses}. Firstly, they rely on simple pointwise distributions, which model one-dimensional variables. Doing so overlooks the direct connections between neural network weights and results in suboptimal initialisations. Secondly, they disregard important architectural information and knowledge from past trainings of similar architectures. This practice amplifies the already substantial financial and environmental costs associated with training networks on large datasets \cite{shang2022one}.\\
In Bayesian Neural Networks (BNN) \cite{blundell2015weight,fuchs2021practical,graves2011practical,jospin2022hands,kingma2015variational,senapati2020bayesian,wenzel2020good}, where neural networks are combined with stochastic modelling, researchers assume arbitrary Gaussian distributions over the weights \cite{silvestro2020prior}. However, the deep-weight-prior \cite{atanov2018deep} stands out as an exception. It employs generative models to learn a \textit{local} distribution of trained weights, which is then utilised for initialising small BNNs. Their findings indicate that leveraging generative models for neural network initialisation leads to improved convergence.\\
Building upon the concept of leveraging generative models for initialisation, we present \textbf{two groups of new initialisation methods}:\textit{local} and \textit{global} initialisations address the limitations of traditional methods. They employ complex weight distributions, incorporate comprehensive architectural information, and leverage past training experiences. While they share the common goal of improving initialisation, they differ in the scope of initialisation, the architectural information considered, and the specific generative models used to learn weight distributions. \\
With \textbf{\textit{local} initialisations}, we sample small groups of weights from learned distributions using variational autoencoders (VAE) \cite{keller2021topographic,kingma2013auto,van2017neural}. In contrast, \textbf{\textit{global} initialisations} obtain full weight sets conditioned on the network's architecture, accomplished through graph hypernetworks (GHN) \cite{knyazev2021parameter}.\\
We evaluate these methods on state-of-the-art deep convolutional neural networks (CNN), focusing on convergence speed, accuracy and ensembling. Our findings reveal that \textit{global} initialisations achieve \textbf{faster initial convergence and higher accuracy}. However, these benefits come with a potential trade-off: \textbf{reduced generalisation ability} of ensembles.\\
To deepen our understanding of this trade-off, we explore the diversity of the different initialisation methods. We identify missing diversity in GHN initialisations as the cause for its reduced generalisation ability. Motivated by this, \textbf{we propose a modification called the }\textbf{Noise GHN}, which introduces diversity into GHNs by modelling a non-deterministic distribution. This is achieved by injecting noise into the GHN decoder and using a modified loss function to encourage the production of diverse weight sets. The noise injection also increases model robustness, enabling better learning of essential features and preventing overfitting. Consequently, the Noise GHN might be able to transfer learned knowledge to different image distributions.\\

\section{Related Work and Preliminaries}
Variational autoencoders and graph hypernetworks are the backbone of our approach. We offer a concise overview of these frameworks, highlighting their essential aspects and functionalities relevant to our work.
\subsubsection{Variational Autoencoders}
VAEs are generative neural networks designed to learn and capture the underlying distribution of a given dataset. Once trained, these networks can be effectively utilised to generate diverse forms of data, such as images \cite{sun2014deep,van2017neural} or natural language \cite{devlin2018bert,radford2018improving}. \\
It consists of two main components: an encoder model $Q_{\boldsymbol\Phi}(\textbf{Z}|\textbf{X})$ and a decoder model $P_{\boldsymbol\theta}(\textbf{X}|\textbf{Z})$. The encoder maps input data to a latent space, while the decoder model reconstructs the input data from the latent space. VAEs are optimised by maximising a lower bound on the log-likelihood of the data. This lower bound is called evidence lower bound (ELBO). Doing so, they learn a meaningful latent space representation that captures the underlying data structure.\\
For a given data point $x$, a prior $P(\textbf{Z})$ on the latent variables $\textbf{Z}$ and the parameters $\boldsymbol \Phi$ and $\boldsymbol \theta$ of the VAE, the ELBO can be expressed as:
\begin{equation}
     \mathcal{L}_{ \boldsymbol \theta, \boldsymbol \Phi}(\textbf{x}) =   \mathbb{E}_{q_{ \boldsymbol \Phi}(\textbf{z}|\textbf{x})}\left[ \log p_{ \boldsymbol \theta}(\textbf{x}|\textbf{z})\right] - \text{KL}[Q_{ \boldsymbol \Phi}(\textbf{Z}|\textbf{x})||P(\textbf{Z})]. \label{eqn:VAE}
\end{equation}
The VAE's optimisation is driven by the expected value of the log density. It promotes the generation of outputs similar to the input. Simultaneously, the KL Divergence between the encoder model and the prior acts as a regularisation term for the latent space. By default, we assume simple Gaussian distributions for the prior, encoder, and decoder models conditioned on the input. This choice offers the advantage of closed-form KL Divergence computation for equation \ref{eqn:VAE}, without significantly limiting the model's expressive capacity.\\\\
Introducing a discrete latent space can encourage the model to learn a more structured representation, as it has to navigate the limited capacity of such a space. To achieve this, Vector Quantisation (VQ) is employed. VAEs utilising VQ are called Vector Quantised Variational Autoencoders (VQVAE) \cite{van2017neural}. There, continuous outputs from the encoder network are mapped to discrete points in the latent space using a codebook. To enhance convergence, a codebook loss is incorporated. The authors also suggest employing an autoregressive model called PixelCNN \cite{van2016conditional} alongside the VQVAE, which we denote as VQVAE*. \\
Although there are various VAE variations based on different encoder and decoder assumptions \cite{burgess2018understanding,keller2021topographic,lee2021meta,simonovsky2018graphvae}, this work does not delve into exploring them. We study the VQVAE's ability to initialise neural networks.
\subsubsection{Graph HyperNetworks}
The GHN \cite{knyazev2021parameter,zhang2018graph} is a generative model that produces complete sets of network weights. Unlike VAEs, which are limited to fixed-size data, a single GHN can be used with various network architectures. The resulting weights are immediately effective, achieving an impressive $58.6\%$ accuracy on CIFAR-10 for a ResNet-50 architecture that it has never seen before. The GHN combines two key concepts: the hypernetwork \cite{ha2016hypernetworks} and the graph network \cite{scarselli2008graph}.\\
A hypernetwork is a neural network that takes input and produces weights for another neural network. The inputs can include various information such as network architecture or data points. The hypernetwork is trained by directly backpropagating the loss of the predicted parameters into the hypernetwork.\\
The second key component is the graph network, which is specifically designed to process graph-based inputs with varying shapes. During the forward pass, a graph network updates the states of the nodes by propagating information along the edges of the graph. By representing network architectures as a computational graphs, the GHN leverages the graph network to obtain states for each node. These states are then used to generate weights for each layer by feeding them into the hypernetwork. Finally, the produced weights are normalised and adjusted to match the dimensions of each layer using slicing and tiling techniques.\\
To optimise the GHN $H_{\boldsymbol \theta}$ with parameters $\boldsymbol \theta$, we employ mini-batch optimisation over two sets of data: batches $b$ of images and batches $b_m$ of architectures:
\begin{equation}
    \mathcal{L} =  \sum_{i=1}^b \sum_{j=1}^{b_m} L\left( f\left( \textbf{x}_i,a_j,H_{\boldsymbol \theta}(a_j)\right),\textbf{y}_i\right),
\end{equation}
where $f^{(k)}(\textbf{x},a,\textbf{w})$ represents the forward-pass of input $\textbf{x}$ through the network architecture $a$ with weights $\textbf{w}$. The function $L$ denotes the loss function used in the optimisation process.\\
Our approach differs from the authors \cite{knyazev2021parameter} work, as they focus on generating deterministic weights with high performance in a single forward pass.
\section{Methods}
VAEs and GHNs capture expressive distributions, enabling us to leverage them for initialising state-of-the-art convolutional classification networks. Their key strength lies in their ability to incorporate architecture information and leverage knowledge gained from past trainings of similar architectures. We now explain further how we utilise these generative models.
\subsection{Local Initialisations}
Our focus on \textit{local} initialisations revolves around capturing patterns within the weights of CNNs. As convolutional networks progress, they transition from extracting edges and colour blobs in early layers to capturing higher-level features in later layers \cite{qin2018convolutional}. Motivated by the specific attributes of filters like edge detectors, we assume that the weights within a CNN layer follow an unknown distribution.\\
Given a neural network architecture, we train a set of variational autoencoders, one for each layer, to learn the unknown local distributions. Once trained, these generative models enable the production of weights that adhere to the learned distributions.\\
We evaluate our \textit{local} initialisations on the well known architecture of the ResNet-20. For every layer, we learn the distribution of the $3 \times 3$ weight slices that constitute the convolutional kernels. To this end, we separately train $100$ ResNet-20s on the training datasets. Afterwards we take the parameters that perform best on the corresponding validation sets. As advised in \cite{atanov2018deep}, we remove the slices with a low $l_2$, by disregarding the slices whose $l_2$ is in the lowest $5\%$ for every layer. We call these datasets of weight-slices the \textit{Weight-Datasets}. Finally, we train VAEs and VQVAEs on it to capture the underlying distributions. 
\subsection{The Global Approach}
We introduce the Noise GHN to transfer the ability to produce instantly performing network weights to ensembles. Furthermore we adapt the training routine of the GHNs to reduce training resources.
\begin{figure}
    \centering
    \includegraphics[width=0.95\textwidth]{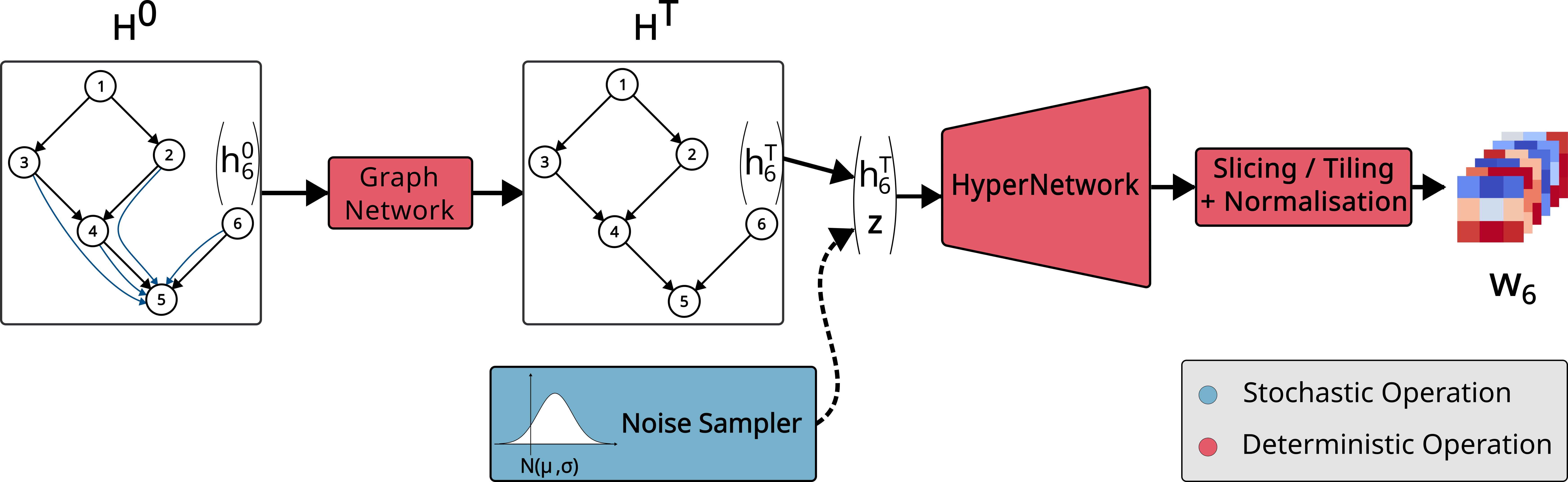}
    \caption{Functioning principle of the Noise GHN. The input, a network architecture, is expressed by a computational graph together with initial hidden states $H^0$ for every node. The graph network propagates information through the computational graph resulting in final hidden states $H^T$, encoding the function of each node. Every final hidden state $h_i^T$ is then fed into a hypernetwork, together with a sampled noise vector. The hypernetwork is encouraged by the loss function to produce performant and diverse weights for every node. In a final step, the produced weights are normalised and fitted to the layer dimensions.}
    \label{fig:noise_ghn}
\end{figure}
\subsubsection{Reducing GHN Training Resources}
The DEEPNETS1M \cite{knyazev2021parameter} dataset trains the GHN, featuring a million network architectures and validation/test sets of $500$/$500$ architectures. Every iteration samples $64$ images and eight architectures. This training setup is conducted over $300$ epochs of the image dataset, which exceeds our computational resources.\\
To conserve resources, we reduce the number of training architectures from eight to three and train for only $30$ instead of $300$ image dataset epochs. To ensure the effectiveness of training, we select training architectures similar to the evaluation architecture of a ResNet-20. Specifically, we choose a ResNet-32, a ResNet-44, and a ResNet-56 and use all three architectures in every iteration.\\
To compensate for the reduced amount of training and training network diversity, we initialise the graph networks of the GHNs by using already trained weights, which are provided by Knyazev \cite{knyazev2021parameter}. As we modify the hypernetwork for the Noise GHN, we initialise all GHN hypernetworks from scratch in order to ensure a fair comparison.
\subsubsection{Modified GHN}
Since the GHN is a deterministic model when given a fixed architecture, its employment results in more similar weights in the trained ensemble members. As ensembles benefit from the diversity of their members, this lack of diversity poses a potential problem.\\
To achieve different weights in every forward pass, we integrate noise into the GHN, as shown in Figure \ref{fig:noise_ghn}. Specifically, we sample and append a noise vector to every final hidden state of the graph network. This design allows the hypernetwork to generate varied outputs, while learning to produce effective weights based on architectures encodings.\\
Additionally, we encourage diversity in the loss function by including a similarity loss into the overall loss function. To measure similarity, we first perform two forward passes of the Noise GHN for the same architecture. Then we calculate the similarity of predictions on a batch of images. Doing so, we ensure that the two produced weight sets correspond to two different functions.\\
Given a Noise GHN $H_{\boldsymbol \theta}$, with parameters $\boldsymbol \theta$, a network architecture $a$, a sample $(\textbf{x},\textbf{y})$ from the dataset $D=\{(\textbf{x}_i,\textbf{y}_i)\}_{i=1}^N$ and samples $\xi_1,\xi_2$ from a noise distribution, the similarity loss $\mathcal{L}_{S}$ is calculated as:
\begin{equation}
\mathcal{L}_{S} \big( \textbf{x},a,H_{\boldsymbol \theta},\xi_1,\xi_2) \big) = \text{CoSim} \Big (f^{(1)} \big( \textbf{x},a,H_{\boldsymbol \theta}(a,\xi_1 \big) ,f^{(2)} \big( \textbf{x},a,H_{\boldsymbol \theta}(a,\xi_2) \big) \Big )
\end{equation}
where $f^{(k)}(\textbf{x},a,\textbf{w})$ represents the $k$-th forward-pass with input $\textbf{x}$ into the network architecture $a$ with weights $\textbf{w}$, $L$ some loss function and CosSim  the \textit{Cosine similarity}. \\
The overall loss function $\mathcal{L}$ for a mini-batch $b$ of images and the three training architectures can then be expressed as:
\begin{equation}
\mathcal{L} =  \sum_{i=1}^b \sum_{j=1}^3 \Bigg [ \sum_{k=1}^2 \bigg [ L\Big( f^{(k)}\big(  \textbf{x}_i,a_j,H_{\boldsymbol \theta}(a_j,\xi_k)\big),\textbf{y}_i\Big) \bigg ] + \mathcal{L}_{S} \big( \textbf{x}_i,a_j,H_{\boldsymbol \theta},\xi_1,\xi_2 \big) \Bigg].\\
\end{equation}
The conceptual differences between the \textit{local} and the \textit{global} initialisations overall can be explained in two dimensions, as seen in Figure \ref{fig:approach}. First, \textit{global} initialisations utilise a distribution which enables initialisation of all weights in a meaningful connected way - as opposed to initialising all weights independently from another. As the GHN is trained to produce already working weights, the weights are already synchronised. On the contrary, the \textit{local} initialisations focus on small weight groups, who are initialised independently from another. Secondly, the \textit{global} initialisations condition the weights on the whole architecture, while the \textit{local} initialisations confine themselves to information about layer position.

\section{Experiments and Results}

We evaluate various network initialisations on CIFAR-10 \cite{krizhevsky2009learning} and the medical PatchCamelyon \cite{veeling2018rotation} (PCam) dataset to assess accuracy and convergence. Additionally, we evaluate ensemble accuracy and calibration on the out-of-distribution (OOD) CIFAR-C \cite{hendrycks2019robustness} dataset. Finally, we investigate the Noise GHN's generalisation ability from natural images to the medical domain. For more detailed information on the datasets and experimental setups, please refer to the supplementary material.
\subsubsection{Convergence Speed}
\begin{figure}[t]
    \centering
    \begin{subfigure}{0.49\linewidth}
        \centering
        \includegraphics[width =\linewidth]{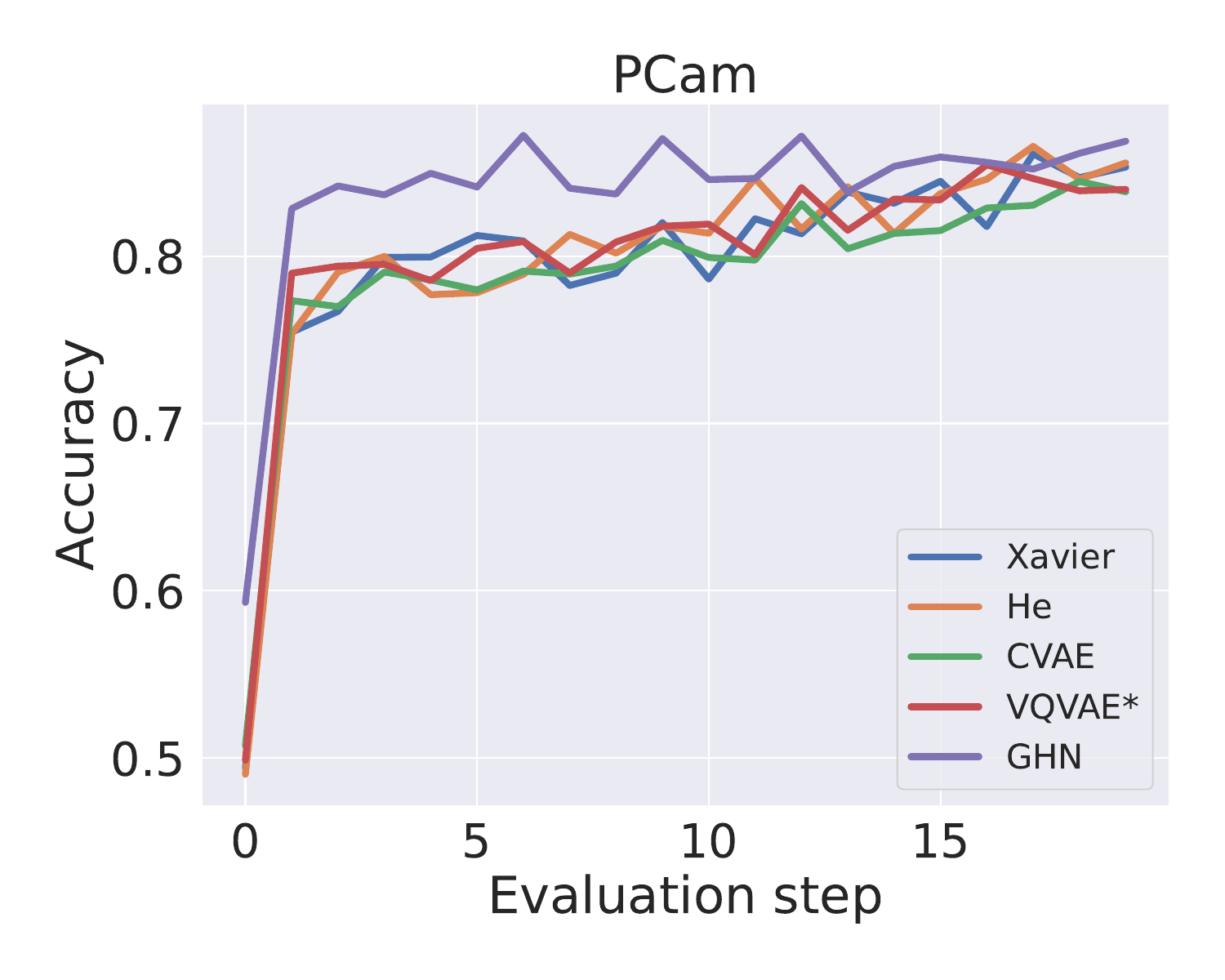}
    \end{subfigure}
    \begin{subfigure}{0.49\linewidth}
        \centering
        \includegraphics[width =\linewidth]{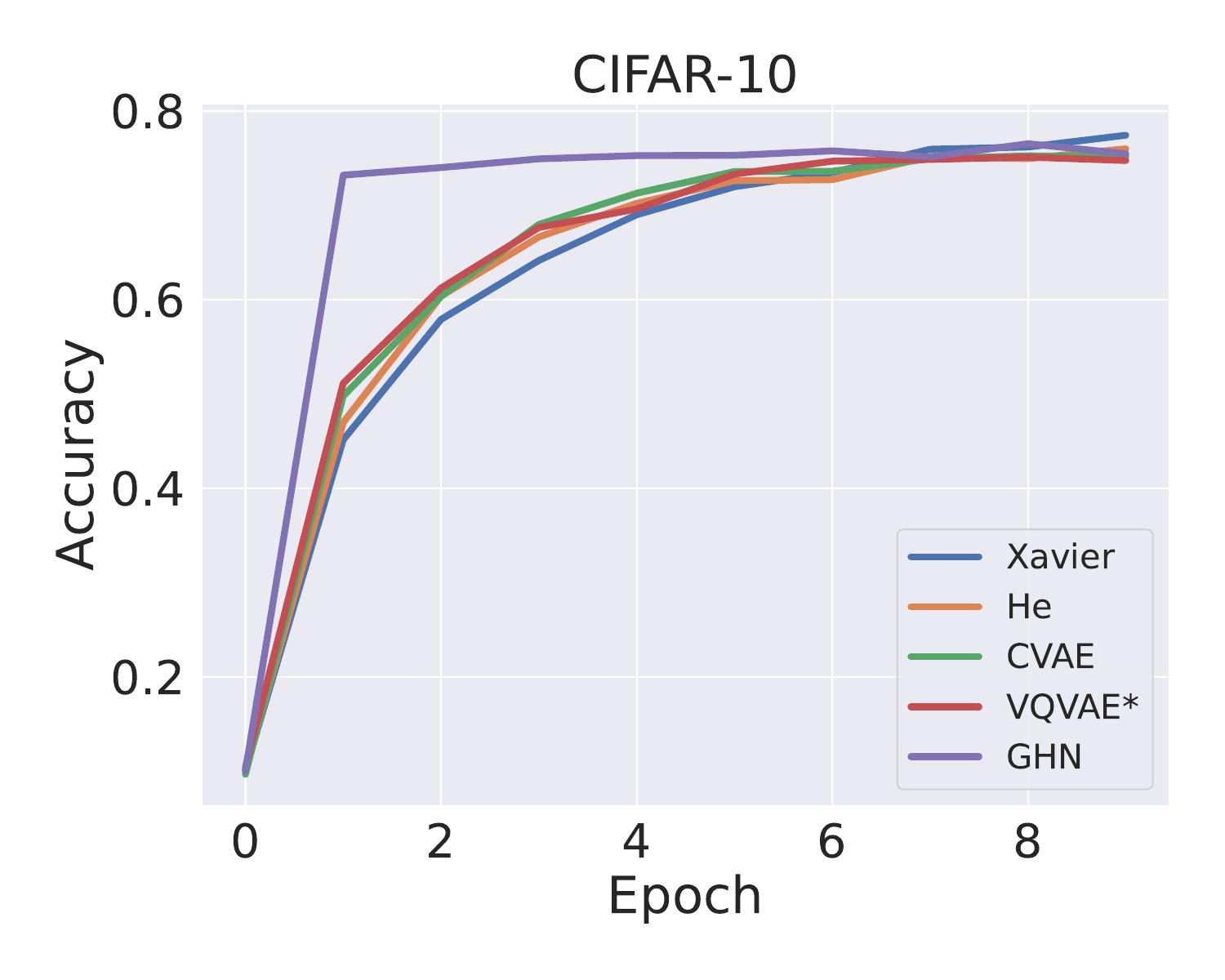}
    \end{subfigure}    
    \caption{Trajectories of validation accuracy during the initial training phase. GHN initialisations outperform other initialisation with a faster initial convergence. }
\label{fig:initial_conv_speed}
\end{figure}

We study convergence speed resulting from different initialisations by evaluating every $300$ batches on PCam and every epoch on the CIFAR-10 dataset. Results are averaged over $25$ trainings per initialisation. Comparing the convergence speed, we measure the steps to reach specified accuracy thresholds. PCam's thresholds are $0.80$ and $0.85$, while CIFAR-10's are $0.65$ and $0.75$. The visualisations can be found in Figure \ref{fig:initial_conv_speed}, while more detailed results, along with an explanation of our threshold selection, are available in Table 1 and Figure 1 in the supplementary material.\\
Our \textit{global} initialisation consistently outperforms other methods in early training. On the PCam dataset, GHN-initialised ResNet-20s reach the first threshold after the initial evaluation step, while other initialisations require five or more steps. The GHN initialisation consistently outperforms others, achieving the second thresholds almost twice as fast.\\
\textbf{The \textit{global} scope} for information and initialisation enhances the GHN's initial convergence, reaching thresholds \textbf{up to 5 times faster}. However, there is no noticeable difference between pointwise standard initialisations and \textit{local} initialisations. Information about the target layer and initialising small weight groups shows no training improvement.\\
Since the learning rate scheduling is the same for all initialisations, we cannot assess how the accelerated initial convergence affects the overall training time. Despite not reaching state-of-the-art accuracies during training due to our training routine, this experiment yields promising results. Therefore, further research in this direction holds great potential.

\subsubsection{Accuracy}
\begin{figure}[t]
    \centering
    \begin{subfigure}{0.49\linewidth}
        \centering
        \includegraphics[width =\linewidth]{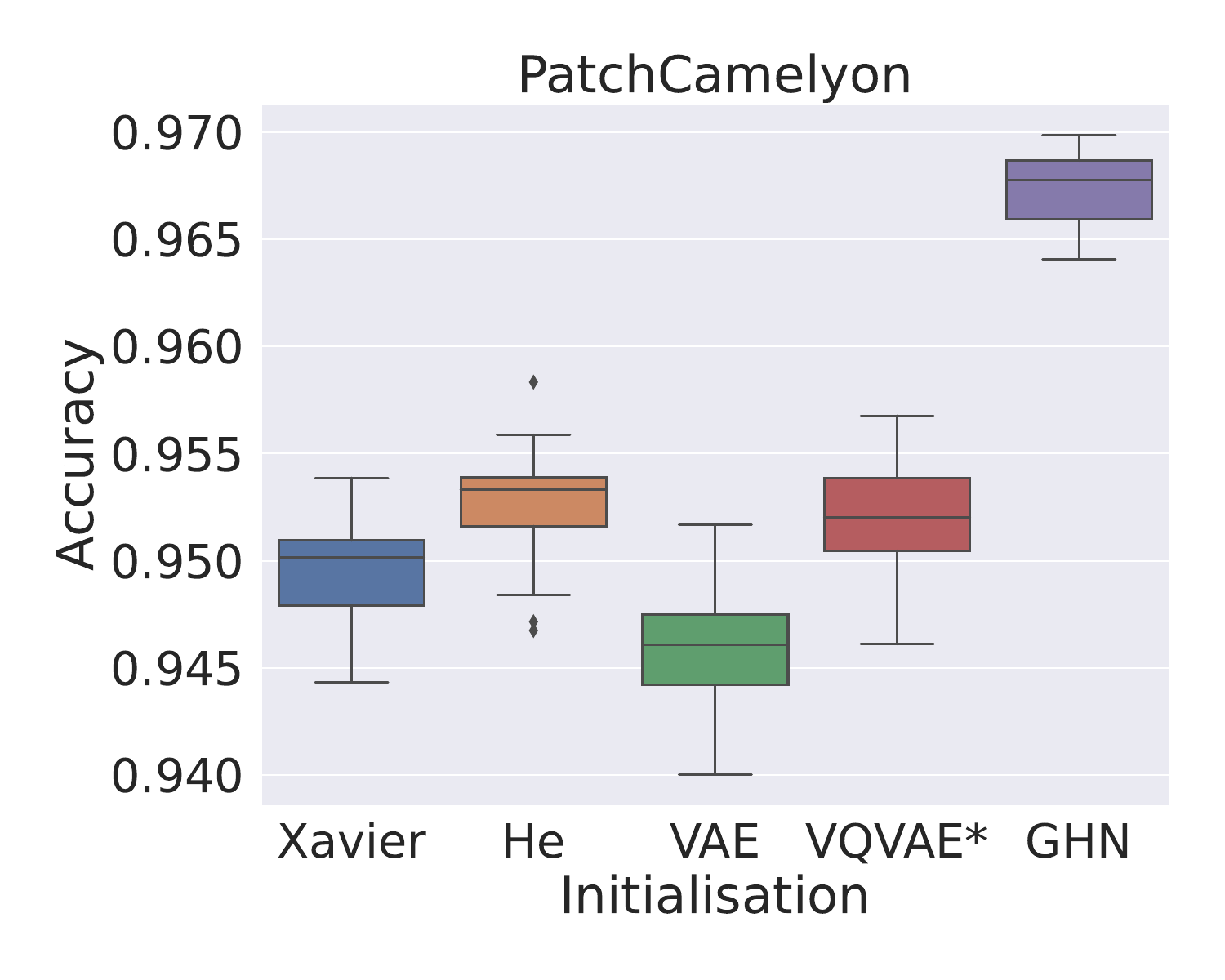}
    \end{subfigure}
    \begin{subfigure}{0.49\linewidth}
        \centering
        \includegraphics[width =\linewidth]{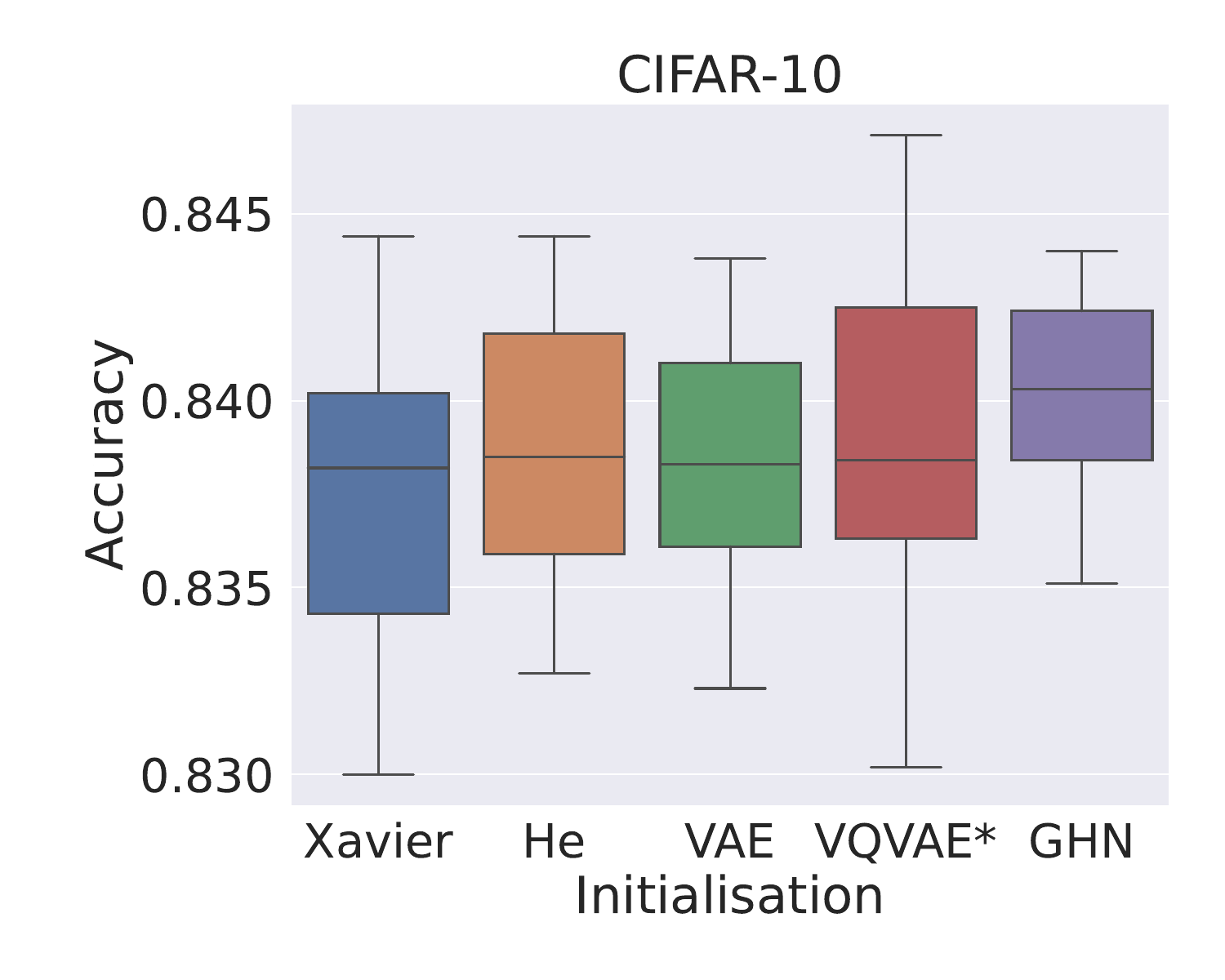}
    \end{subfigure}    
    \caption{Boxplots displaying the resulting test set accuracies of 25 ResNet-20s per initialisation. The GHN results the highest median accuracy on both datasets, especially on the PCam dataset where all $25$ GHN initialisations outperform every other initialisation. }
\label{fig:test_acc_pcam}
\end{figure}
We assess test set accuracies of ResNet-20s with different initialisations, training 25 networks for each. Figure \ref{fig:test_acc_pcam} shows the resulting accuracies. The \textit{Global} initialisation on PCam outperforms others, with \textbf{all GHN initialisations} achieving \textbf{higher accuracy}. On CIFAR-10, GHN initialisation achieves a higher median accuracy but by a smaller margin. However, no notable differences appear between traditional and \textit{local} initialisations.\\
The \textit{global} scope for information and initialisation leads to faster convergence and higher test set accuracy. Conversely, there are no significant performance differences between \textit{local} and standard initialisations. The payoff doesn't increase proportionally with expanded scope and information. This can be explained by a key advantage of our global GHN initialisation, which produces instantly performing and synchronised weight sets.\
The importance of this \textbf{synchronicity for fast initial convergence and higher accuracy} is a crucial insight, highlighting the potential benefits of employing generative models for initialisations.

\subsubsection{OOD Ensembling}
Accurate initialisations and fast convergence are crucial in the context of ensembling. Ensembles provide higher accuracy and improved uncertainty estimation. These advantages are particularly important for OOD data. To examine the effect of different initialisations on ensembles, we evaluate their expected calibration error (ECE) and accuracy. Note that a lower ECE indicates better calibration. We provide a detailed explanation of its calculation in the supplemental material.\\
The ensembles consist of the $25$ trained ResNet-20 networks per initialisation.
We sample $20$ ensembles with $5$ members each and calculate the ECE across all 5 corruption levels of the CIFAR-C dataset. We present the results in Figure \ref{fig:ece_ens_cifarc}. The corresponding accuracy results can be found in Figure $2$ in the supplementary material, showing similar trends.\\
The VAE and VQVAE* based initialisations yield the lowest ECE values. Surprisingly, the \textit{global} GHN initialisation shows the highest ECE value, despite its faster convergence and higher accuracy. High accuracy does not necessarily guarantee good OOD performance. Nevertheless, the magnitude of the performance shift is surprising. This discrepancy is further investigated in the next section.
\begin{figure*}[h!]
    \centering
    \includegraphics[width=\textwidth]{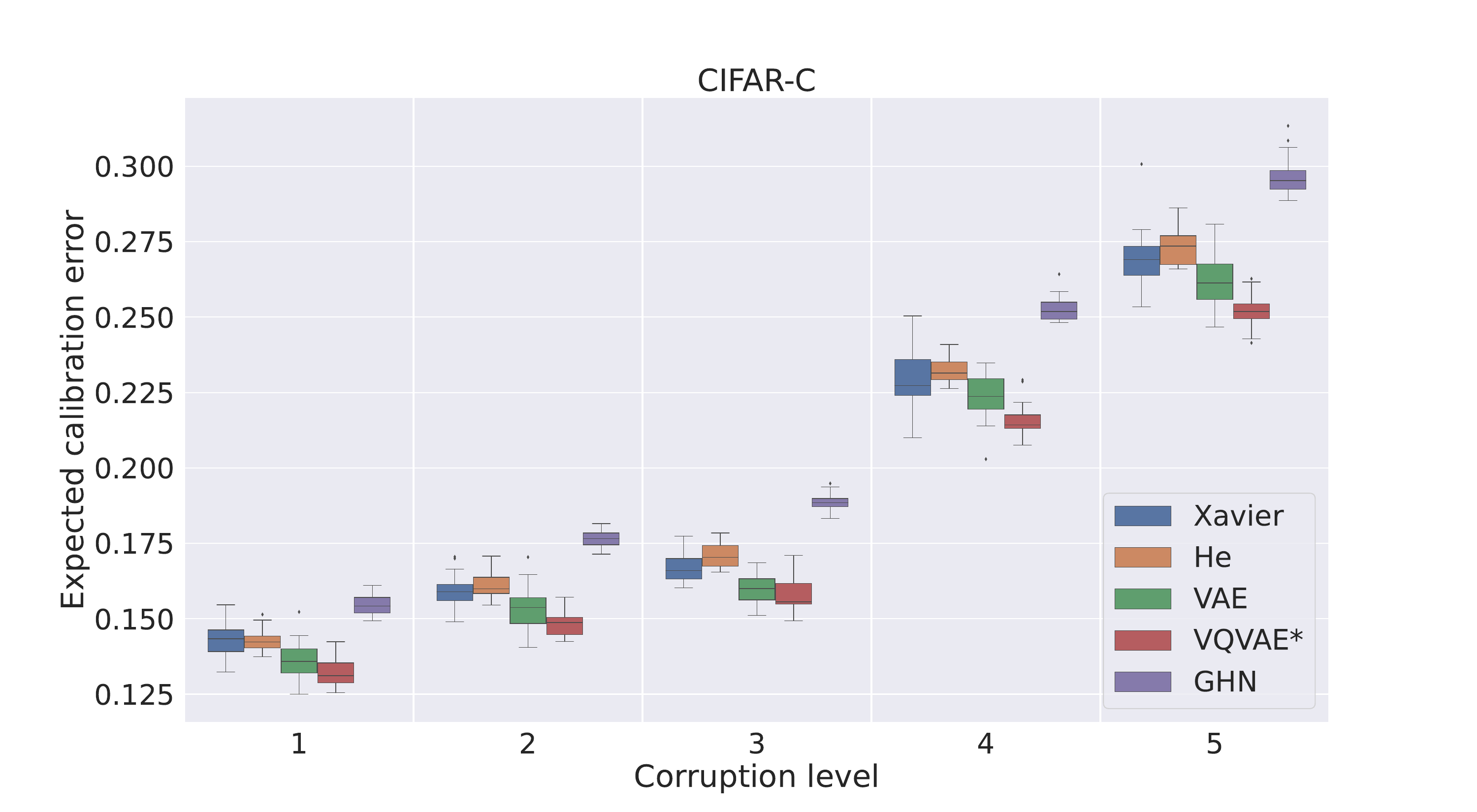}
    \caption{Boxplots of the resulting ECE of 20 ensembles consisting of each five members on the OOD CIFAR-C dataset.}
    \label{fig:ece_ens_cifarc}
\end{figure*}
\newlength{\mylen}
\setlength{\mylen}{1.97cm}
\begin{figure*}[t]
\centering
    \includegraphics[height=\mylen]{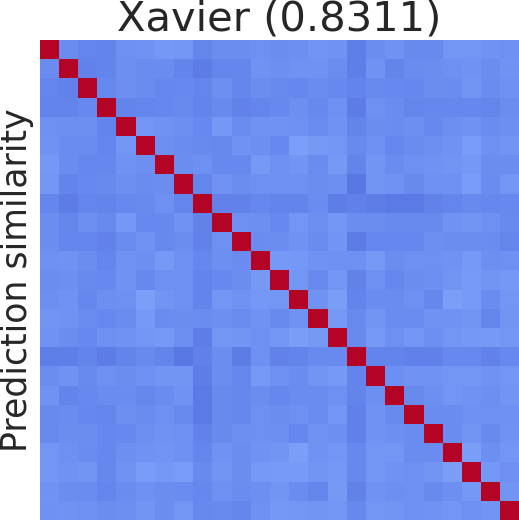}
    \includegraphics[height=\mylen]{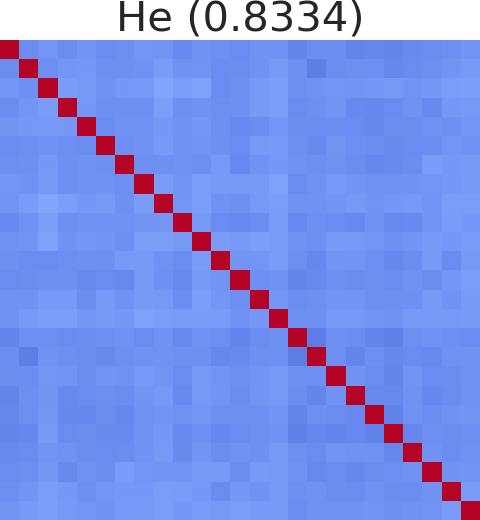}
    \includegraphics[height=\mylen]{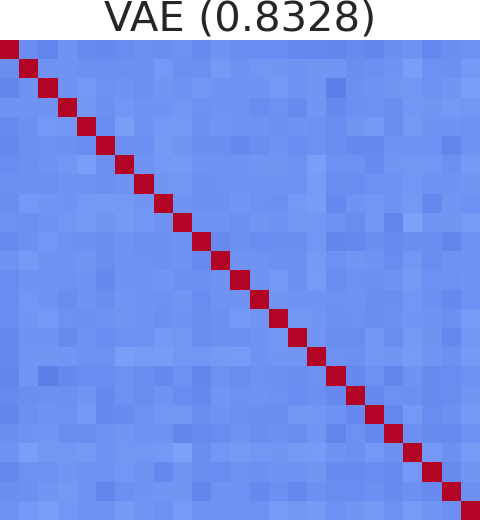}
    \includegraphics[height=\mylen]{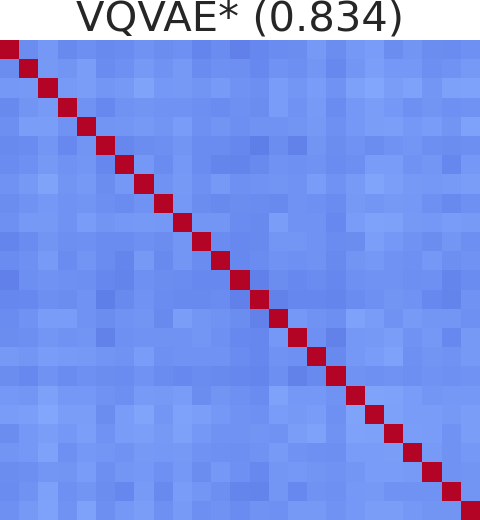}
    \includegraphics[height=\mylen]{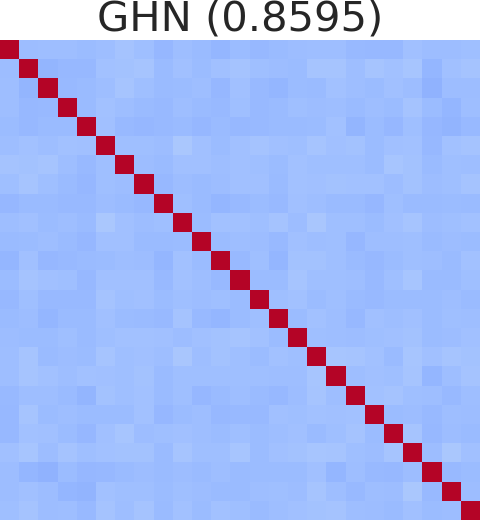}
    \includegraphics[height=\mylen]{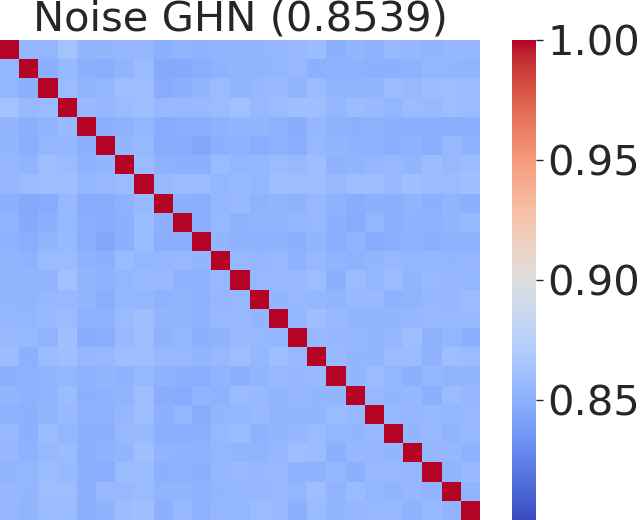}

    \includegraphics[height=\mylen]{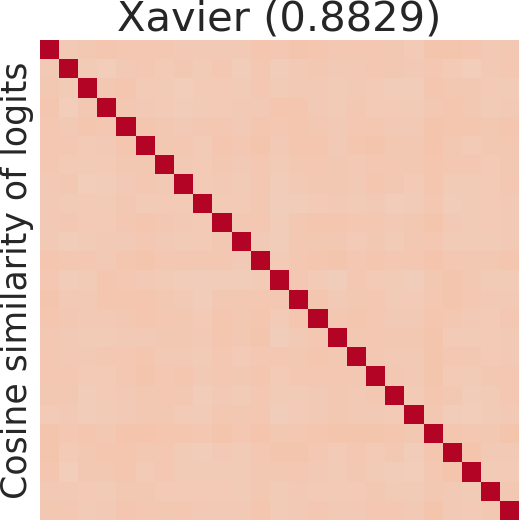}
    \includegraphics[height=\mylen]{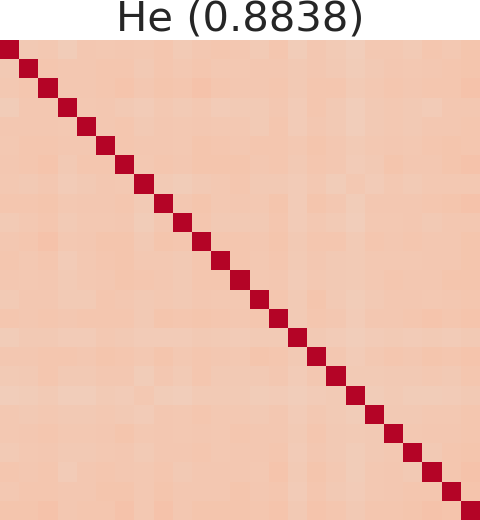}
    \includegraphics[height=\mylen]{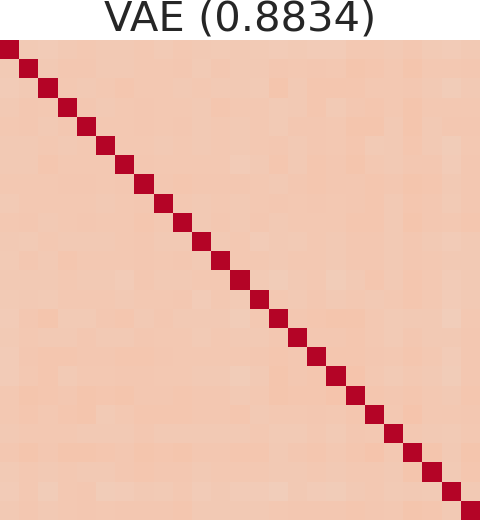}
    \includegraphics[height=\mylen]{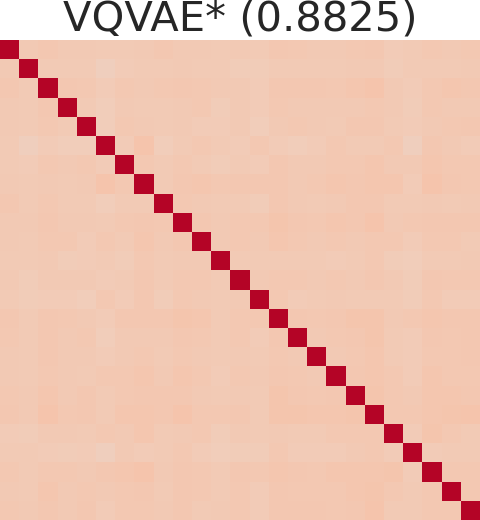}
    \includegraphics[height=\mylen]{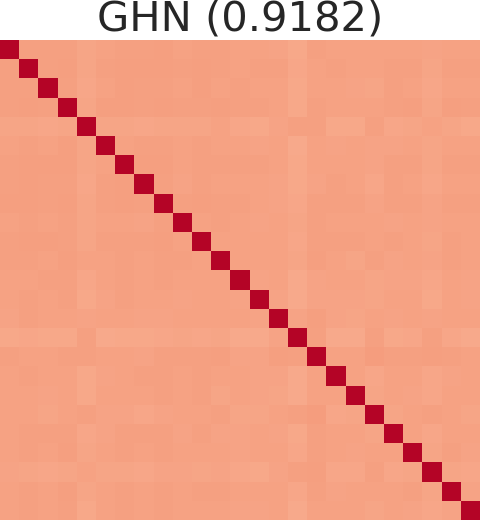}
    \includegraphics[height=\mylen]{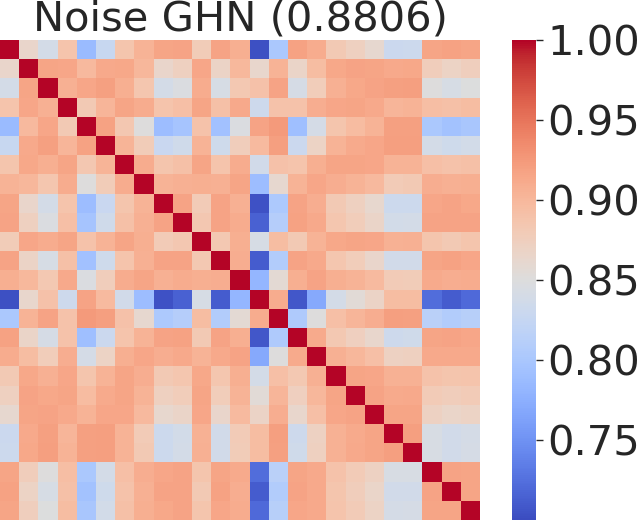}
   
    \caption{Pairwise similarities of trained weights for different initialisations. Lower similarity score (dark blue) is better, as it implies a higher diversity. The GHN obtains the highest similarity measures, indicating missing diversity. The Noise GHN performs better than the GHN in terms of both similarity measures and all other initialisations in terms of their cosine similarity.}
    \label{fig:cifar_pred_sim}
\end{figure*}

\subsubsection{Prediction Similarities}
Ensemble uncertainty estimation and accuracy rely on member diversity \cite{fort2019deep}. We analyse this diversity by examining prediction similarity and cosine similarity of logits on the CIFAR-10 test set. Figure \ref{fig:cifar_pred_sim} illustrates pairwise similarities, and average scores, computed over the strict upper diagonal matrix. Lower similarity scores are preferable, as they indicate higher diversity.\\
\textbf{Standard} and \textbf{\textit{local}} initialisations exhibit similar diversities in both measures, while \textbf{GHN} initialisations show significantly lower diversity. The GHN produces the same set of weights for a fixed architecture, which results in more similar trained weights, reducing diversity.\\
To address this, our Noise GHN introduces diversification. We input a sampled vector into the Noise GHN decoder, and the modified loss function encourages low cosine similarity logits, ensuring diverse weights in each forward pass.\\
The effect is evident on the right-hand side of Figure \ref{fig:cifar_pred_sim}. The Noise GHN initialisation yields the lowest cosine similarity and slightly improves on the GHN in prediction similarity. Notably, only the Noise GHN initialisation achieves cosine similarity scores below $0.8$.\\
\noindent The Noise GHN significantly enhances diversity, leading to a clear impact on the OOD ECE, as shown in Figure \ref{fig:ece_noiseghn}. It consistently outperforms the GHN across all five corruption levels.\\
Furthermore, the Noise GHN retains the advantages of global initialisation, surpassing the convergence speed achieved by GHN initialisation, as seen in Table 1 in the supplementary material. This improvement is due to the sampled noise, which enhances weight robustness, akin to techniques like Monte-Carlo dropout.
\subsubsection{Knowledge Transfer}
\begin{figure}[t]
\begin{subfigure}{0.45\textwidth}
    \centering
    \includegraphics[width=\linewidth]{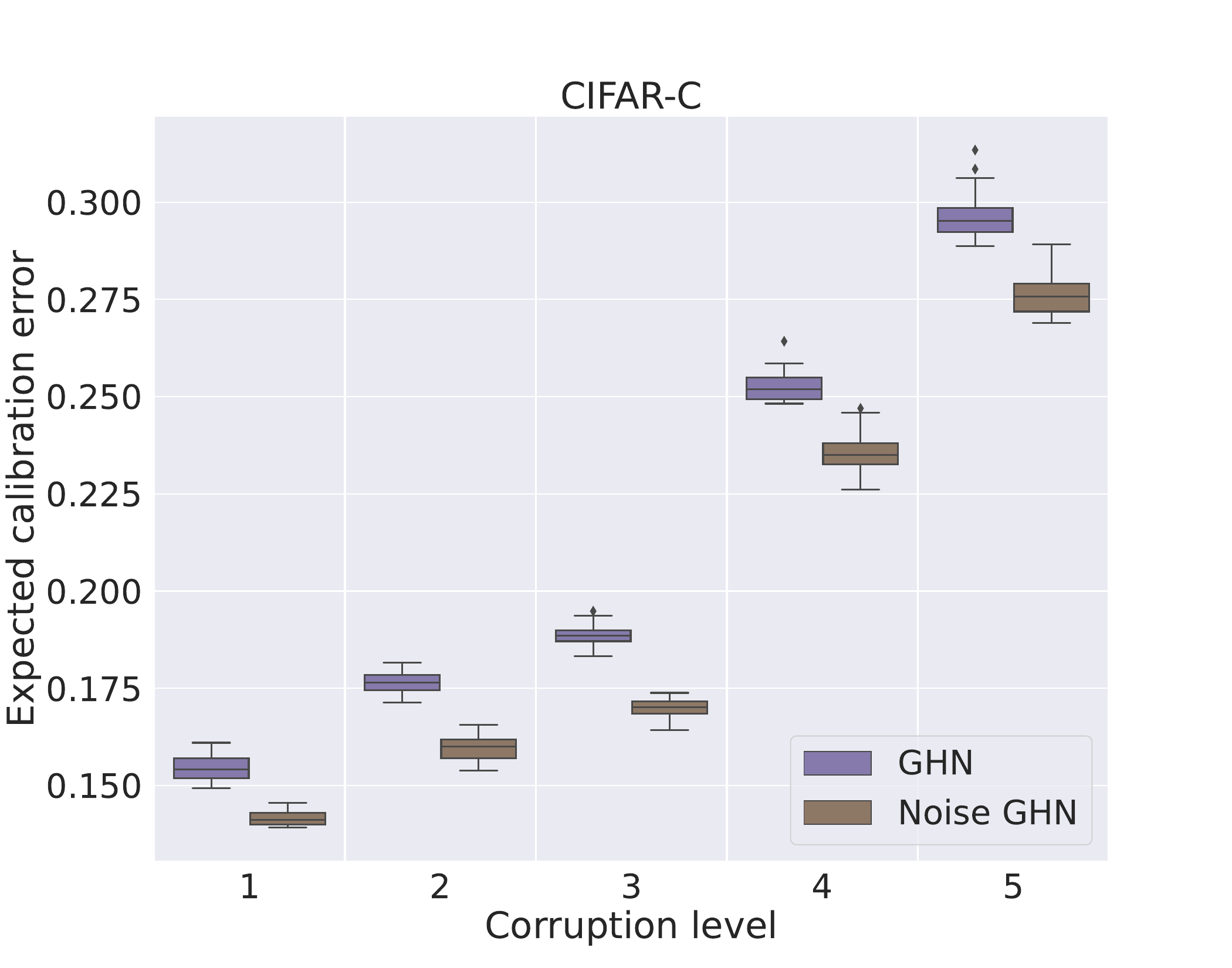}
\end{subfigure}
\hspace{0.01\textwidth}
\begin{subfigure}{0.54\textwidth}
    \centering
    \includegraphics[width=\linewidth]{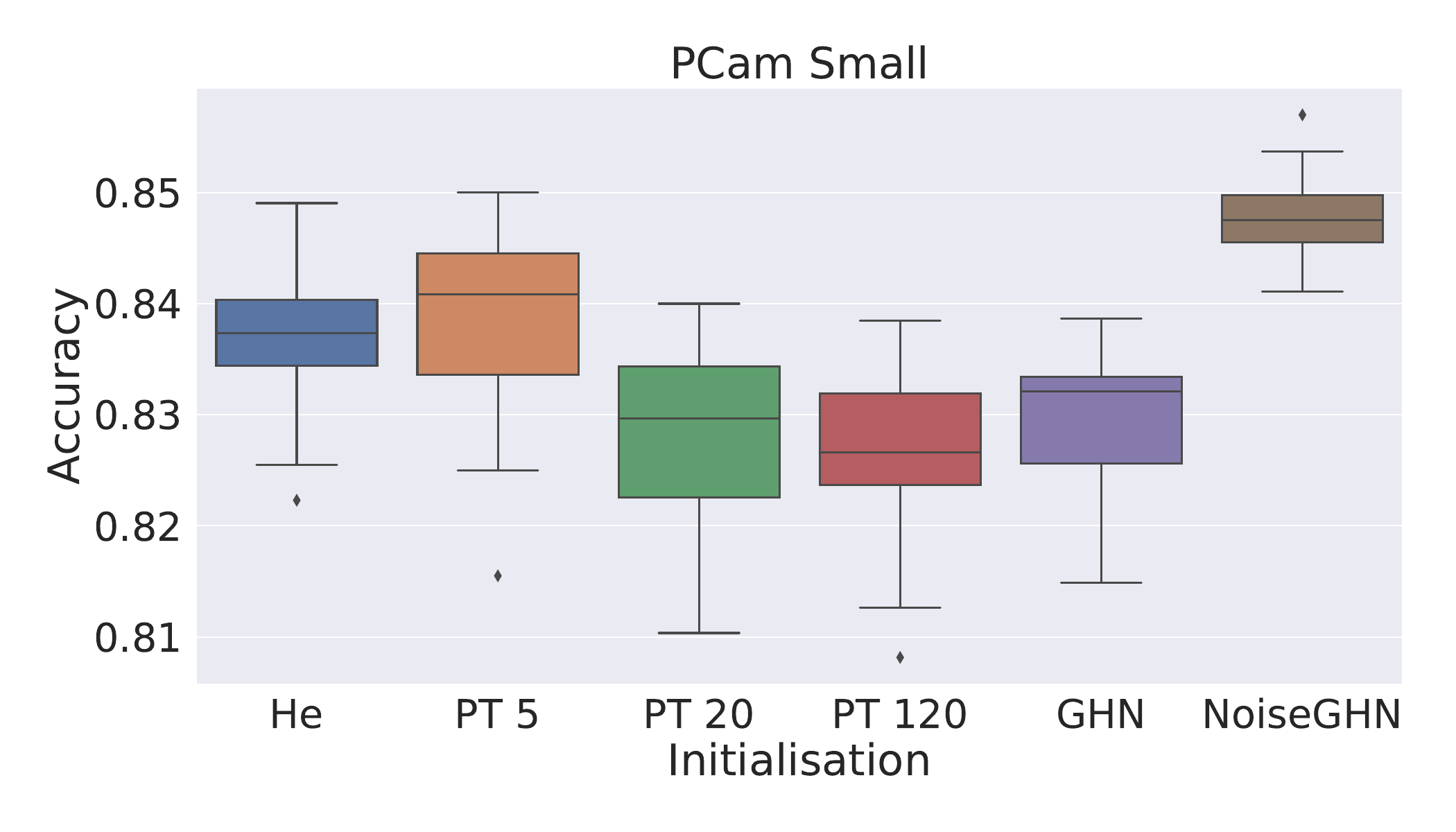}
    \vspace{1px}
\end{subfigure}
\caption{Comparison of the GHN and Noise GHN. Left: Due to the diversity of the Noise GHN it improves upon the GHN over all levels of corruption. Right: The Noise GHN is able to transfer relevant knowledge from CIFAR-10 to the PCam Small dataset and outperform all other initialisations. We abbreviate pretraining on CIFAR-10 for five epochs as "PT 5" and similarly for other epochs. }
\label{fig:ece_noiseghn}
\end{figure}

We explore the ability of our methods to transfer knowledge between image distributions, particularly between CIFAR-10's natural images and PCam's medical data. Such knowledge transfer can be especially beneficial for low-data tasks. To this end, we consider a dataset of $1000$ training patches from PCam, with additional $400$ validation patches for model selection and evaluate on a test dataset of a separate $10.000$ patches. We call this downscaled dataset PCam Small. All splits underlie a strict $50\%/50\%$ split of tumorous and benign patches. After initialisation the ResNet-20s are trained for $40$ epochs using our standard procedure. To account for the different dataset sizes, we adapt the learning rate schedule: After $20$ epochs, the learning rate is halved, and after $30$ epochs, the learning rate is reduced by a factor of five.\\
We compare the results to two types of baselines: He initialisation of a ResNet-20 and pretraining on CIFAR-10 for different amounts of epochs. We compare the performance of these baselines to a GHN and Noise GHN, who have been trained on the CIFAR-10 dataset.\\
Figure \ref{fig:ece_noiseghn} displays the results, showing overall weaker performance compared to the training on PCam due to the smaller dataset and distribution shift. The Noise GHN outperforms other initialisations, supported by a significant t-test result (p$<0.0001$) against all other initialisations. This result can be attributed to the noise injection, making it more robust and preventing overfitting. This experiment indicates Noise GHN's potential for knowledge transfer between different image distributions. For a comparison of all initialisation methods in this work, please refer to Figure 3 in the supplementary material.\\
Our results align with Knyazev's findings \cite{knyazev2021parameter}, showing GHN parameters can be fine-tuned on small, different datasets. However, their experiments include a less severe distribution shift. Our experiment's weaker GHN performance might be attributed to the changes we made in the training setup. Nevertheless, further research is required to fully understand the reasons behind it.
\subsubsection{Limitations}
Our work is focused on evaluating different generative models with regard to their potential for the initialisation of networks. As a first step, we evaluate these new initialisation methods on in-distribution data. In comparison to the training of a single network, this setup of training and employing on the same dataset is very costly. However, the potential savings when applying our methods to various datasets would justify these costs.\\
Due to the immense computational power required for the training of the GHN by Knyazev \cite{knyazev2021parameter}, we downscaled the training setup. Thus, a meaningful comparison to the original network is not possible.\\
We train and evaluate our methods on two in-distribution datasets, one OOD dataset and one architecture. Additional experiments are needed to verify, if the advantages extend to different datasets and network architectures.\\
Currently, the Noise GHN falls short of the CIFAR-C ensemble performance of \textit{local} and standard initialisations. Nevertheless, we believe future large-scale experiments can further unlock its potential.

\section{Conclusion}
We explore incorporating additional knowledge into initialisations. Our study demonstrates that initialisations based on learned weight distributions with a \textbf{\textit{global} scope can offer significant advantages}.  We show these advantages in terms of convergence speed and accuracy for feed-forward convolutional neural networks. We identify the key factor for the benefits of the global initialisations to be the \textbf{synchronicity of the produced weights}. In contrast, \textit{local} initialisations are missing synchronous produced weights. \\
This work also reveals that deterministic \textit{global} initialisations result in weaker OOD ensembling accuracy and calibration. However, the \textbf{introduction of the Noise GHN improves diversity and performance}. Further generalisation of our approach is especially intriguing due to the potential environmental and economic cost savings. We provide the groundwork for utilising \textbf{\textit{global} initialisations in the area of ensembling} and show that the Noise GHN might be able to transfer learned knowledge to different image distributions. We showcase their strengths on single networks and introduce non-deterministic global initialisations.
\newpage
\bibliographystyle{splncs04}
\bibliography{bibliography}

\begin{thebibliography}{10}
\providecommand{\url}[1]{\texttt{#1}}
\providecommand{\urlprefix}{URL }
\providecommand{\doi}[1]{https://doi.org/#1}

\bibitem{atanov2018deep}
Atanov, A., Ashukha, A., Struminsky, K., Vetrov, D., Welling, M.: The deep
  weight prior. arXiv preprint arXiv:1810.06943  (2018)

\bibitem{blundell2015weight}
Blundell, C., Cornebise, J., Kavukcuoglu, K., Wierstra, D.: Weight uncertainty
  in neural network. In: International conference on machine learning. pp.
  1613--1622. PMLR (2015)

\bibitem{burgess2018understanding}
Burgess, C.P., Higgins, I., Pal, A., Matthey, L., Watters, N., Desjardins, G.,
  Lerchner, A.: Understanding disentangling in $\beta$-vae. arXiv preprint
  arXiv:1804.03599  (2018)

\bibitem{cao2023swin}
Cao, H., Wang, Y., Chen, J., Jiang, D., Zhang, X., Tian, Q., Wang, M.:
  Swin-unet: Unet-like pure transformer for medical image segmentation. In:
  Computer Vision--ECCV 2022 Workshops: Tel Aviv, Israel, October 23--27, 2022,
  Proceedings, Part III. pp. 205--218. Springer (2023)

\bibitem{ciregan2012multi}
Ciregan, D., Meier, U., Schmidhuber, J.: Multi-column deep neural networks for
  image classification. In: 2012 IEEE conference on computer vision and pattern
  recognition. pp. 3642--3649. IEEE (2012)

\bibitem{clevert2015fast}
Clevert, D.A., Unterthiner, T., Hochreiter, S.: Fast and accurate deep network
  learning by exponential linear units (elus). arXiv preprint arXiv:1511.07289
  (2015)

\bibitem{devlin2018bert}
Devlin, J., Chang, M.W., Lee, K., Toutanova, K.: Bert: Pre-training of deep
  bidirectional transformers for language understanding. arXiv preprint
  arXiv:1810.04805  (2018)

\bibitem{fort2019deep}
Fort, S., Hu, H., Lakshminarayanan, B.: Deep ensembles: A loss landscape
  perspective. arXiv preprint arXiv:1912.02757  (2019)

\bibitem{fuchs2021practical}
Fuchs, M., Gonzalez, C., Mukhopadhyay, A.: Practical uncertainty quantification
  for brain tumor segmentation. In: Medical Imaging with Deep Learning (2021)

\bibitem{girshick2014rich}
Girshick, R., Donahue, J., Darrell, T., Malik, J.: Rich feature hierarchies for
  accurate object detection and semantic segmentation. In: Proceedings of the
  IEEE conference on computer vision and pattern recognition. pp. 580--587
  (2014)

\bibitem{glorot2010understanding}
Glorot, X., Bengio, Y.: Understanding the difficulty of training deep
  feedforward neural networks. In: Proceedings of the thirteenth international
  conference on artificial intelligence and statistics. pp. 249--256. JMLR
  Workshop and Conference Proceedings (2010)

\bibitem{gou2021knowledge}
Gou, J., Yu, B., Maybank, S.J., Tao, D.: Knowledge distillation: A survey.
  International Journal of Computer Vision  \textbf{129},  1789--1819 (2021)

\bibitem{graves2011practical}
Graves, A.: Practical variational inference for neural networks. In: Advances
  in neural information processing systems. pp. 2348--2356 (2011)

\bibitem{ha2016hypernetworks}
Ha, D., Dai, A., Le, Q.V.: Hypernetworks. arXiv preprint arXiv:1609.09106
  (2016)

\bibitem{he2015delving}
He, K., Zhang, X., Ren, S., Sun, J.: Delving deep into rectifiers: Surpassing
  human-level performance on imagenet classification. In: Proceedings of the
  IEEE international conference on computer vision. pp. 1026--1034 (2015)

\bibitem{he2016deep}
He, K., Zhang, X., Ren, S., Sun, J.: Deep residual learning for image
  recognition. In: Proceedings of the IEEE conference on computer vision and
  pattern recognition. pp. 770--778 (2016)

\bibitem{hendrycks2019robustness}
Hendrycks, D., Dietterich, T.: Benchmarking neural network robustness to common
  corruptions and perturbations. Proceedings of the International Conference on
  Learning Representations  (2019)

\bibitem{huang2020unet}
Huang, H., Lin, L., Tong, R., Hu, H., Zhang, Q., Iwamoto, Y., Han, X., Chen,
  Y.W., Wu, J.: Unet 3+: A full-scale connected unet for medical image
  segmentation. In: ICASSP 2020-2020 IEEE International Conference on
  Acoustics, Speech and Signal Processing (ICASSP). pp. 1055--1059. IEEE (2020)

\bibitem{jing2020self}
Jing, L., Tian, Y.: Self-supervised visual feature learning with deep neural
  networks: A survey. IEEE transactions on pattern analysis and machine
  intelligence  \textbf{43}(11),  4037--4058 (2020)

\bibitem{jospin2022hands}
Jospin, L.V., Laga, H., Boussaid, F., Buntine, W., Bennamoun, M.: Hands-on
  bayesian neural networks—a tutorial for deep learning users. IEEE
  Computational Intelligence Magazine  \textbf{17}(2),  29--48 (2022)

\bibitem{keller2021topographic}
Keller, T.A., Welling, M.: Topographic vaes learn equivariant capsules.
  Advances in Neural Information Processing Systems  \textbf{34},  28585--28597
  (2021)

\bibitem{kingma2014adam}
Kingma, D.P., Ba, J.: Adam: A method for stochastic optimization. arXiv
  preprint arXiv:1412.6980  (2014)

\bibitem{kingma2013auto}
Kingma, D.P., Welling, M.: Auto-encoding variational bayes. ICLR  (2014)

\bibitem{kingma2015variational}
Kingma, D.P., Salimans, T., Welling, M.: Variational dropout and the local
  reparameterization trick. Advances in neural information processing systems
  \textbf{28} (2015)

\bibitem{knyazev2021parameter}
Knyazev, B., Drozdzal, M., Taylor, G.W., Romero, A.: Parameter prediction for
  unseen deep architectures. In: Thirty-Fifth Conference on Neural Information
  Processing Systems (2021)

\bibitem{krizhevsky2009learning}
Krizhevsky, A., Hinton, G., et~al.: Learning multiple layers of features from
  tiny images  (2009)

\bibitem{lee2021meta}
Lee, D.B., Min, D., Lee, S., Hwang, S.J.: Meta-gmvae: Mixture of gaussian vae
  for unsupervised meta-learning. In: International Conference on Learning
  Representations (2021)

\bibitem{cuda}
NVIDIA, Vingelmann, P., Fitzek, F.H.: Cuda, release: 10.2.89 (2020),
  \url{https://developer.nvidia.com/cuda-toolkit}

\bibitem{van2016conditional}
Van~den Oord, A., Kalchbrenner, N., Espeholt, L., Vinyals, O., Graves, A.,
  et~al.: Conditional image generation with pixelcnn decoders. Advances in
  neural information processing systems  \textbf{29} (2016)

\bibitem{ovadia2019can}
Ovadia, Y., Fertig, E., Ren, J., Nado, Z., Sculley, D., Nowozin, S., Dillon,
  J., Lakshminarayanan, B., Snoek, J.: Can you trust your model's uncertainty?
  evaluating predictive uncertainty under dataset shift. Advances in neural
  information processing systems  \textbf{32} (2019)

\bibitem{pytorch}
Paszke, A., Gross, S., Massa, F., Lerer, A., Bradbury, J., Chanan, G., Killeen,
  T., Lin, Z., Gimelshein, N., Antiga, L., Desmaison, A., Kopf, A., Yang, E.,
  DeVito, Z., Raison, M., Tejani, A., Chilamkurthy, S., Steiner, B., Fang, L.,
  Bai, J., Chintala, S.: Pytorch: An imperative style, high-performance deep
  learning library. In: Wallach, H., Larochelle, H., Beygelzimer, A.,
  d\textquotesingle Alch\'{e}-Buc, F., Fox, E., Garnett, R. (eds.) Advances in
  Neural Information Processing Systems 32, pp. 8024--8035. Curran Associates,
  Inc. (2019),
  \url{http://papers.neurips.cc/paper/9015-pytorch-an-imperative-style-high-performance-deep-learning-library.pdf}

\bibitem{qin2018convolutional}
Qin, Z., Yu, F., Liu, C., Chen, X.: How convolutional neural networks see the
  world---a survey of convolutional neural network visualization methods.
  Mathematical Foundations of Computing  \textbf{1}(2) (2018)

\bibitem{radford2018improving}
Radford, A., Narasimhan, K., Salimans, T., Sutskever, I., et~al.: Improving
  language understanding by generative pre-training  (2018)

\bibitem{ren2015faster}
Ren, S., He, K., Girshick, R., Sun, J.: Faster r-cnn: Towards real-time object
  detection with region proposal networks. Advances in neural information
  processing systems  \textbf{28} (2015)

\bibitem{russakovsky2015imagenet}
Russakovsky, O., Deng, J., Su, H., Krause, J., Satheesh, S., Ma, S., Huang, Z.,
  Karpathy, A., Khosla, A., Bernstein, M., et~al.: Imagenet large scale visual
  recognition challenge. International journal of computer vision
  \textbf{115}(3),  211--252 (2015)

\bibitem{scarselli2008graph}
Scarselli, F., Gori, M., Tsoi, A.C., Hagenbuchner, M., Monfardini, G.: The
  graph neural network model. IEEE transactions on neural networks
  \textbf{20}(1),  61--80 (2008)

\bibitem{senapati2020bayesian}
Senapati, J., Roy, A.G., P{\"o}lsterl, S., Gutmann, D., Gatidis, S., Schlett,
  C., Peters, A., Bamberg, F., Wachinger, C.: Bayesian neural networks for
  uncertainty estimation of imaging biomarkers. In: International Workshop on
  Machine Learning in Medical Imaging. pp. 270--280. Springer (2020)

\bibitem{shang2022one}
Shang, F., Yang, Y., Yang, D., Wu, J., Wang, X., Xu, Y.: One hyper-initializer
  for all network architectures in medical image analysis. arXiv preprint
  arXiv:2206.03661  (2022)

\bibitem{silvestro2020prior}
Silvestro, D., Andermann, T.: Prior choice affects ability of bayesian neural
  networks to identify unknowns. arXiv preprint arXiv:2005.04987  (2020)

\bibitem{simonovsky2018graphvae}
Simonovsky, M., Komodakis, N.: Graphvae: Towards generation of small graphs
  using variational autoencoders. In: Artificial Neural Networks and Machine
  Learning--ICANN 2018: 27th International Conference on Artificial Neural
  Networks, Rhodes, Greece, October 4-7, 2018, Proceedings, Part I 27. pp.
  412--422. Springer (2018)

\bibitem{strudel2021segmenter}
Strudel, R., Garcia, R., Laptev, I., Schmid, C.: Segmenter: Transformer for
  semantic segmentation. In: Proceedings of the IEEE/CVF international
  conference on computer vision. pp. 7262--7272 (2021)

\bibitem{sun2014deep}
Sun, Y., Chen, Y., Wang, X., Tang, X.: Deep learning face representation by
  joint identification-verification. Advances in neural information processing
  systems  \textbf{27} (2014)

\bibitem{sutskever2013importance}
Sutskever, I., Martens, J., Dahl, G., Hinton, G.: On the importance of
  initialization and momentum in deep learning. In: International conference on
  machine learning. pp. 1139--1147. PMLR (2013)

\bibitem{tan2018survey}
Tan, C., Sun, F., Kong, T., Zhang, W., Yang, C., Liu, C.: A survey on deep
  transfer learning. In: Artificial Neural Networks and Machine Learning--ICANN
  2018: 27th International Conference on Artificial Neural Networks, Rhodes,
  Greece, October 4-7, 2018, Proceedings, Part III 27. pp. 270--279. Springer
  (2018)

\bibitem{van2017neural}
Van Den~Oord, A., Vinyals, O., et~al.: Neural discrete representation learning.
  Advances in neural information processing systems  \textbf{30} (2017)

\bibitem{veeling2018rotation}
Veeling, B.S., Linmans, J., Winkens, J., Cohen, T., Welling, M.: Rotation
  equivariant cnns for digital pathology. In: International Conference on
  Medical image computing and computer-assisted intervention. pp. 210--218.
  Springer (2018)

\bibitem{wan2013regularization}
Wan, L., Zeiler, M., Zhang, S., Le~Cun, Y., Fergus, R.: Regularization of
  neural networks using dropconnect. In: International conference on machine
  learning. pp. 1058--1066. PMLR (2013)

\bibitem{wenzel2020good}
Wenzel, F., Roth, K., Veeling, B., Swiatkowski, J., Tran, L., Mandt, S., Snoek,
  J., Salimans, T., Jenatton, R., Nowozin, S.: How good is the bayes posterior
  in deep neural networks really? In: International Conference on Machine
  Learning. pp. 10248--10259. PMLR (2020)

\bibitem{zhang2018graph}
Zhang, C., Ren, M., Urtasun, R.: Graph hypernetworks for neural architecture
  search. In: 7th International Conference on Learning Representations, ICLR
  2019 (2019)

\end{thebibliography}
\newpage
\appendix
\section{Architectures and Training}
\subsubsection{Datasets and Experimental Setup}
We train the initialisation methods on two image classification datasets: CIFAR-10 \cite{krizhevsky2009learning} and PCam \cite{veeling2018rotation}. The PCam dataset consists of $327,680$ patches sized $96 \times 96$ with a resolution of $0.243$ microns per pixel. These patches contain hematoxylin-eosin (H$\&$E) stained lymph node sections, and each patch is labeled for the presence of metastatic tissue. We use a modified version of the PCam dataset from kaggle, which contains $220,025$ non-duplicate patches. For evaluating OOD performance, we utilise the CIFAR-C \cite{hendrycks2019robustness} dataset.\\
The PCam dataset is split into $70\%/10\%/20\%$, while the CIFAR-10 dataset follows a split of $75\%/8.\Bar{3}\%/16.\Bar{6}\%$ for train, validation and test sets.
\subsubsection{Training the evaluation architecture ResNet-20}
We evaluate the different initialisation approaches on the ResNet-20 architecture since it is well known and generally well-performing. Except for the initialisations, all these network trainings were be conducted using the same hyper-parameters, which are further elaborated in the following. For each initialisation method, including the well-known He and Xavier initialisations, we train $25$ models on each dataset. This amounts to $150$ trained ResNet-20s per dataset.\\
Following \cite{he2016deep}, we train the networks using the Stochastic Gradient Descent (SGD) implementation of PyTorch \cite{pytorch} with a weight decay of $10^{-4}$, a momentum of $0.9$. We use the He-initialisation \cite{he2015delving} and do not employ augmentations. The models are trained with a batch size of $128$ on one GPU for $120$ epochs on CIFAR-10 and $10$ epochs on PatchCamelyon. The learning rate is initiated as $0.1$ and divided by five after $80$ epochs or $7,200$ batches and by two after $100$ epochs or $8,400$ batches on CIFAR-10 and PatchCamelyon, respectively.\\
\subsubsection{Training the local approach}
Motivated by the interpretation of the weight slices as images, we use convolutional encoder-decoder network pairs. To obtain the models hyperparameters, we tune them on the validation split of the Weight-Dataset of the first layer of the ResNet-20 from CIFAR-10 and select the models with the highest validation accuracy. \\
We build the VAE following the $3 \times 3$ encoder and decoder structure provided in Atanov's \cite{atanov2018deep} code. The hidden layer dimension is $32$, while the latent space is $5$-dimensional. The VQVAE* encoder consists of three $3 \times 3$ convolutions with a padding of two. We use the ELU \cite{clevert2015fast} activation after the first two convolutions. For vector quantisation we have nine input vectors with dimension four and a codebook of size $128$. The VQVAE* decoder consists of a $3 \times 3$ convolution, followed by an ELU activation and two $3 \times 3$ transposed convolutions. Both the VQVAE* encoder and decoder have a hidden dimension of $16$\\
Following the original papers \cite{atanov2018deep,van2017neural}, we initialise the two networks using PyTorch default initialisations. Both networks train using a batch size of $128$ and a linearly decreasing learning rate. We train the CVAE with Adam \cite{kingma2014adam}, an initial learning rate of $0.01$ and a weight decay of $1$. Similarly, we train the VQVAE using Adam with an initial learning rate of $0.01$ and a weight decay of $ 10^{-5}$. We use a commitment cost of $\beta=0.25$ and the default Vector Quantisation updates of the authors.
\subsubsection{Training the global approach} 
As described before, we initialise the graph networks of the GHNs using weights, that have been trained on CIFAR-10 and are provided by the authors \cite{knyazev2021parameter}. However we train the hypernetworks of the GHNs from scratch.\\
Like Knyazev \cite{knyazev2021parameter}, we initialise the hypernetworks using PyTorch default initialisations. We train the GHN and its modified versions using Adam for $30$, respectively $6$ epochs on the CIFAR-10 and PatchCamelyon datasets using an image batch size of $b = 64$. The initial learning rate is $10^{-3}$ and multiplied with $0.1$ after $15$ and $20$ epochs on CIFAR-10, respectively $4$ and $5$ epochs on the PatchCamelyon dataset. We use the GHN architecture from \cite{knyazev2021parameter}. To feed noise into the Hypernetwork of the Noise GHN we sample from an $8$-dimensional normal distribution with a diagonal covariance matrix. For this we increase the initial layers size of the Noise GHN's Hypernetwork by $8$.
\subsubsection{Implementation details}
All experiments have been implemented in Python using the PyTorch framework of version $1.10$. The experiments were conducted on a single NVIDIA GeForce GTX 1080 TI and a Tesla T4 using CUDA 11.3 \cite{cuda}. After acceptance we will publish the code on GitHub.  
\section{Experiments and Results}
\subsubsection{The Expected Calibration Error}
Uncertainty estimation is an essential property of neural networks and ensembles, especially in high-stakes applications such as in the medical domain. It reflects the model's ability to asses how confident it should be about a prediction. Multiple ways of measuring the calibration of a model are known. For further discussion we refer to \cite{ovadia2019can}. In this work we use the \textit{Expected Calibration Error} (ECE).\\
Given a dataset $\{\textbf{x}_i,y_i\}_{i=1}^n$ it is computed as the average gap between within bucket accuracy and within bucket predicted probability. For this purpose $s$ buckets $B_i = \{j \in  1 ..n: p_{\boldsymbol \theta}(y_n|\textbf{x}_i) \in (\rho_i,\rho_{i+1}]\}$ are defined, with equidistant points $\rho_i$ in the interval $(0,1]$ and where $y_n$ denotes the ground-truth label. With $\hat{y}_n$ denoting the predicted label and the operator of the squared brackets being $1$ when the expression is true and $0$ else, we can define the accuracy and confidence of a bin as 
\begin{equation*}
    \text{acc}(B_i) = |B_i|^{-1} \sum_{j \in B_i}[y_j = \hat{y}_j] \hspace{1cm} \text{conf}(B_i) = |B_i|^{-1} \sum_{j \in B_i} p_{\boldsymbol \theta}(y_j|\textbf{x}_j).
\end{equation*}
Then we can compute the ECE as
\begin{equation*}
    \text{ECE} = \sum_{i=1}^s \frac{|B_i|}{n} | \text{acc}(B_i) -  \text{conf}(B_i)|,
\end{equation*}
which is the weighted average of the difference between the confidence and accuracy of the bins, measuring the calibration of the predictions. We chose a number of $s=10$ buckets.
\subsubsection{Boxplots}
We present our results using boxplots, where the box represents the three quartiles of the data, with the median depicted as a line in the middle. The whiskers extend to points within $1.5$ interquartile ranges of the lower and upper quartiles. Any observations that fall outside this range are shown as individual outliers.
\subsubsection{Choosing thresholds for the convergence speed}
We compare the convergence speed of the initialisations by measuring how long it takes the model to reach given thresholds. During regular training on the PCam dataset, the original ResNet-20s quickly reach $0.80$ accuracy. Accuracy plateaus at $0.85$ before the first learning rate step. These thresholds were chosen for PCam. Similarly, thresholds of $0.65$ and $0.75$ were chosen for the CIFAR-10 dataset
\newpage
\section{Further Results}
\begin{table}
 \caption{Number of steps after which accuracy thresholds are reached.}
    \label{tab:conv_traj}
            \centering
            \vspace{0.1cm}
            \begin{tabular}{l|c|c}
                \multicolumn{1}{c|}{\textbf{PCam}} & \multicolumn{2}{c}{Eval steps to reach} \\ \hline
Initialisation                     & \multicolumn{1}{c|}{ \hspace*{3mm} 0.80 \hspace*{3mm}}     & \hspace*{3mm} 0.85  \hspace*{3mm}  \\ \hline
                 Xavier   & $5$    &$17$\\
                 He & $7$ &$17$\\
                 CVAE & $9$&$20$\\
                 VQVAE* & $5$&$16$\\
                 GHN & $\textbf{1}$&$6$\\
                 Noise GHN & $\textbf{1}$&$\textbf{3}$
            \end{tabular}
        \hspace{0.05\linewidth}
            \centering
            \begin{tabular}{l|c|c}
                \multicolumn{1}{c|}{\textbf{CIFAR-10}} & \multicolumn{2}{c}{Epochs to reach} \\ \hline
Initialisation                     & \multicolumn{1}{c|}{ \hspace*{2.5mm} 0.65 \hspace*{2.5mm}}     & \hspace*{2.5mm} 0.75  \hspace*{2.5mm}  \\ \hline
                 Xavier   &$ 4 $   &$7$\\
                 He &$ 3$ &$7$\\
                 VAE & $3$&$8$\\
                 VQVAE* & $3$&$8$\\
                 GHN & $\textbf{1}$&$4$\\
                 Noise GHN & $\textbf{1}$&$\textbf{3}$
            \end{tabular}
\end{table}
\begin{figure}
    \centering
    \begin{subfigure}{0.49\linewidth}
        \centering
        \includegraphics[width =\linewidth]{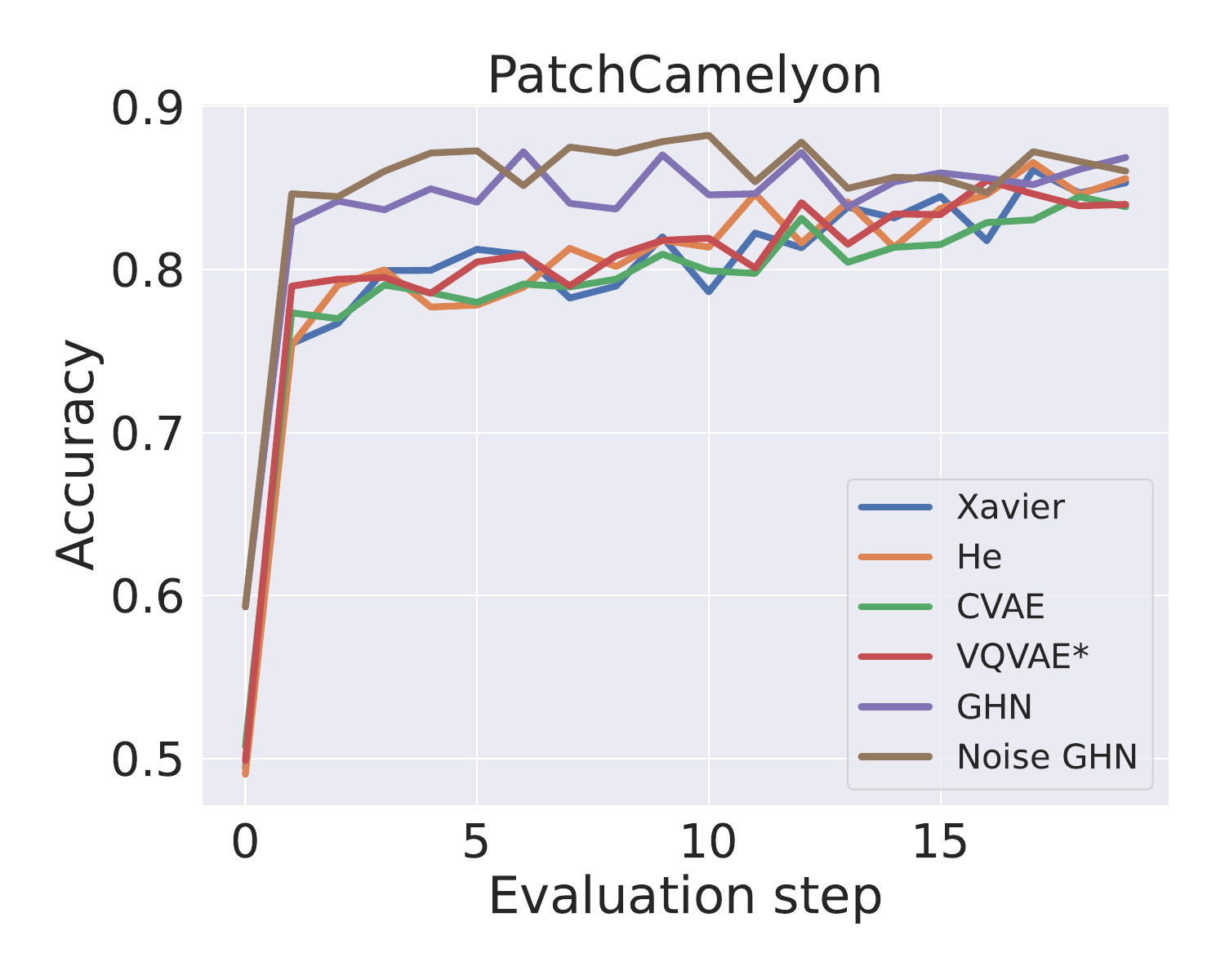}
    \end{subfigure}
    \begin{subfigure}{0.49\linewidth}
        \centering
        \includegraphics[width =\linewidth]{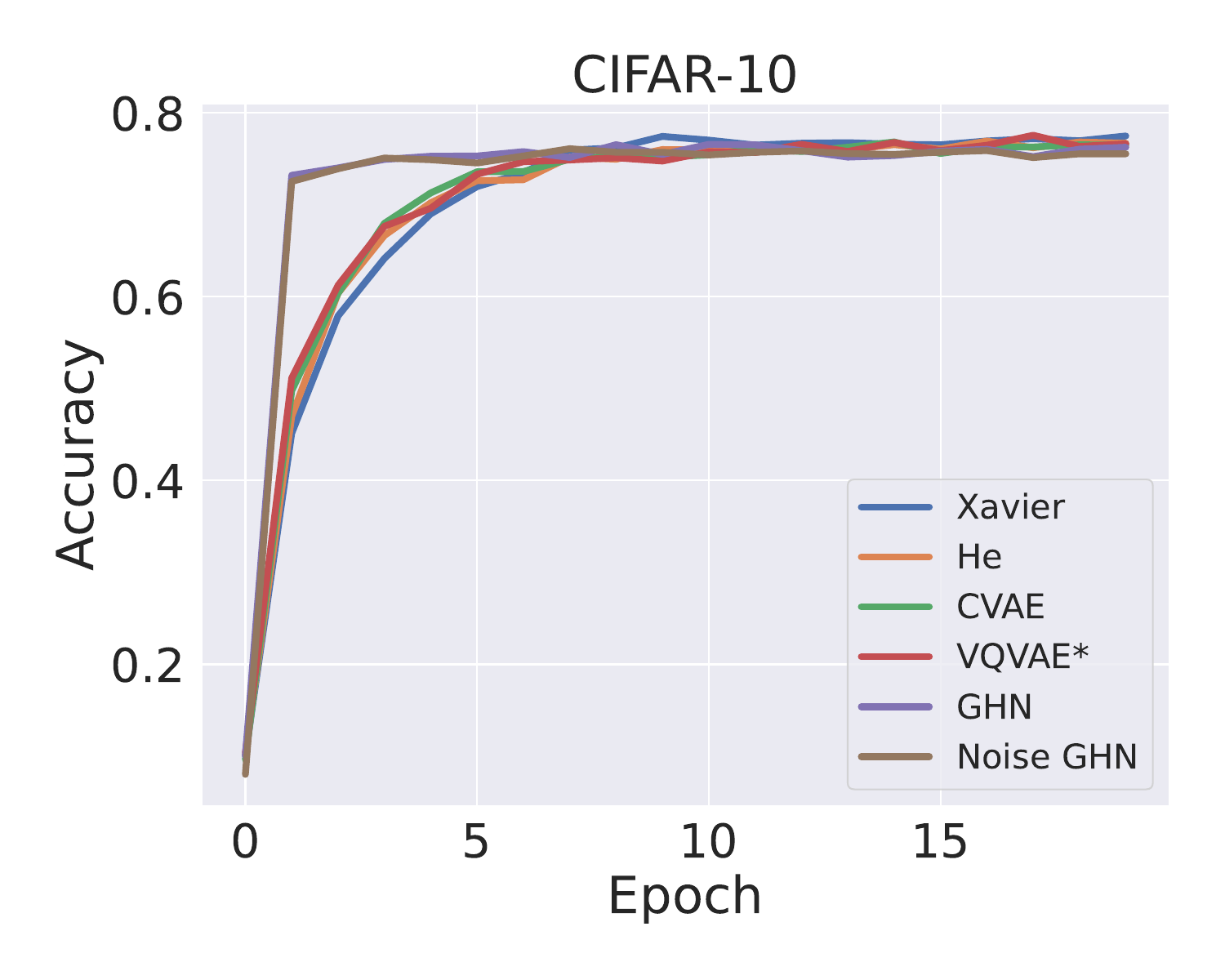}
    \end{subfigure}    
    \caption{Convergence trajectories of the validation accuracy for the different initialisation methods on the PCam and CIFAR-10 dataset. }
\label{fig:val_conv}
\end{figure}
\begin{figure}
    \centering
    \includegraphics[width=\linewidth]{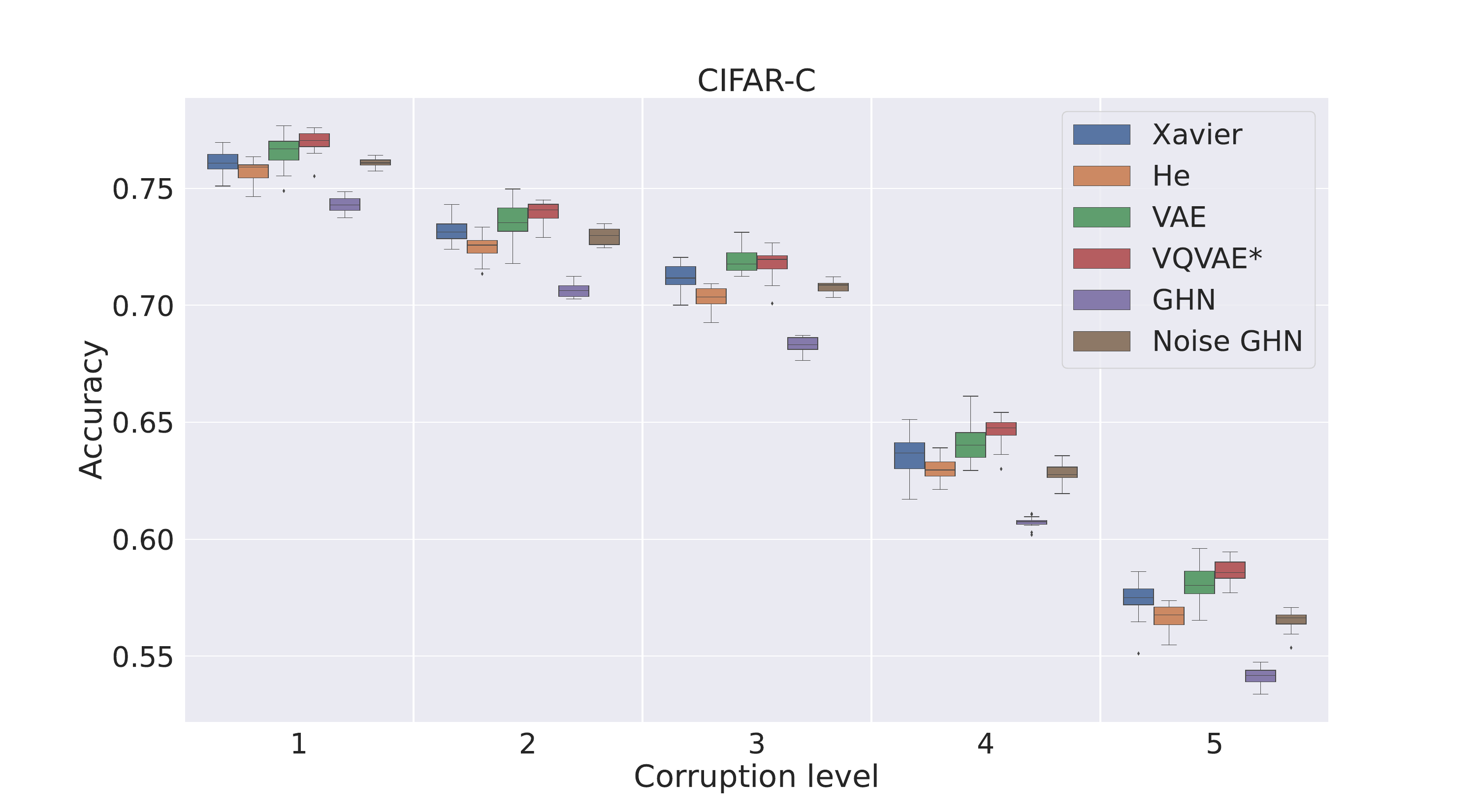}
    \caption{Accuracy boxplots of 20 ensembles consisting of five members each on the OOD CIFAR-C dataset.}
    \label{fig:acc_ens_cifarc}
\end{figure}
\begin{figure}
    \centering
    \includegraphics[width=\linewidth]{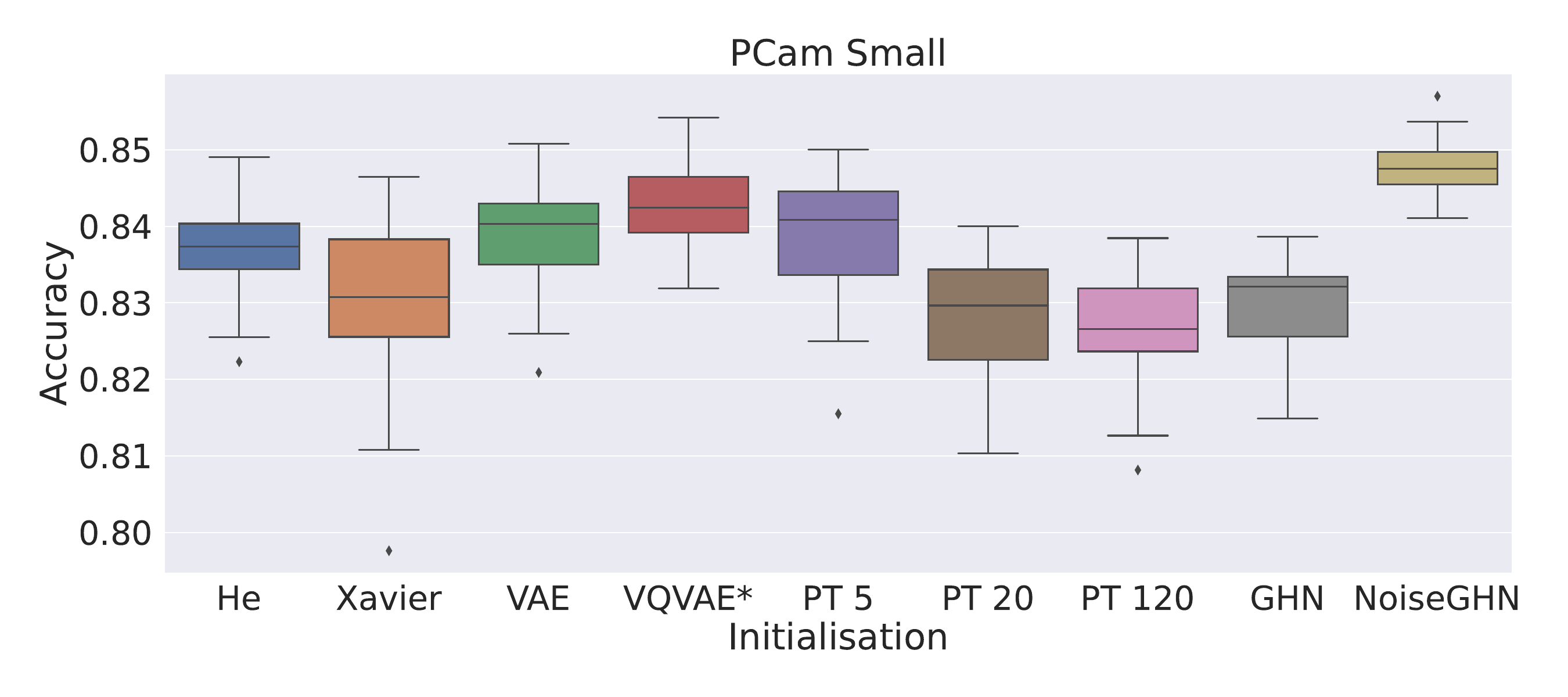}
    \caption{Full comparison of all initialisation used throughout the paper in terms of their ability to transfer relevant knowledge from CIFAR-10 to the PCam Small dataset. The Noise GHN is able to outperform all other initialisations. We abbreviate pretraining on CIFAR-10 for five epochs as "PT 5" and similarly for other epochs.}
    \label{fig:enter-label}
\end{figure}
\clearpage

\end{document}